\documentclass[preprint,12pt,authoryear]{elsarticle}

\usepackage{amssymb,amsmath,amsfonts,bm}
\usepackage{color}
\usepackage{graphicx}
\usepackage{caption}
\usepackage{subcaption}
\usepackage{multirow}
\usepackage{float} 
\usepackage{fullpage}
\usepackage[usenames,dvipsnames]{xcolor}
\usepackage[hidelinks]{hyperref}
\usepackage{booktabs}
\usepackage{placeins}
\usepackage{numprint}

\hypersetup{colorlinks,bookmarksopen,linkcolor=blue,citecolor=blue,urlcolor=blue}

\journal{Elsevier}

\begin{document}

\begin{frontmatter}

\title{Auto-Regressive U-Net for Full-Field Prediction of Shrinkage-Induced Damage in Concrete}

\author[inst1]{Liya Gaynutdinova}

\affiliation[inst1]{organization={Czech Technical University in Prague},
            addressline={Thákurova 7}, 
            city={Prague},
            postcode={16629}, 
            country={Czech Republic}}

\affiliation[inst2]{organization={Eindhoven University of Technology},
            addressline={5600 MB}, 
            city={Eindhoven},
            postcode={P.O. Box 513},
            country={The Netherlands}}

\author[inst1]{Petr Havl\' asek}
\author[inst2]{Ond\v rej Roko\v s  \corref{cor1}}
\author[inst2]{Fleur Hendriks}
\author[inst1]{Martin Do\v sk\' a\v r}

\cortext[cor1]{Corresponding author, email: o.rokos@tue.nl}

\begin{abstract}
This paper introduces a deep learning approach for predicting time-dependent full-field damage in concrete. The study uses an auto-regressive U-Net model to predict the evolution of the scalar damage field in a unit cell given microstructural geometry and evolution of an imposed shrinkage profile. By sequentially using the predicted damage output as input for subsequent predictions, the model facilitates the continuous assessment of damage progression. Complementarily, a convolutional neural network (CNN) utilises the damage estimations to forecast key mechanical properties, including observed shrinkage and residual stiffness. The proposed dual-network architecture demonstrates high computational efficiency and robust predictive performance on the synthesised datasets. The approach reduces the computational load traditionally associated with full-field damage evaluations and is used to gain insights into the relationship between aggregate properties, such as 
shape, size, and distribution, and the effective shrinkage and reduction in stiffness. Ultimately, this can help to optimize concrete mix designs, leading to improved durability and reduced internal damage.
\end{abstract}


\end{frontmatter}

%% \linenumbers

\section{Introduction}
\label{sec:intro}

Shrinkage in concrete is a major concern in construction and civil engineering due to its potential to induce stresses and cracks that compromise durability and may lead to premature structural failures \citep{videla2008guide}. Among the various types of shrinkage, drying shrinkage—caused by moisture loss from the concrete matrix—is typically the most significant \citep{Theiner2017}.

This phenomenon primarily occurs in the mortar phase, as the aggregates are generally considered dimensionally stable. Consequently, the mortar-to-aggregate volume ratio has a substantial influence on the extent of drying shrinkage \citep{BISSONNETTE19991655, GRASSL201085, ZHANG2013500,bisschop-2002}. In addition to this ratio, aggregate characteristics such as particle size, shape, and distribution also affect shrinkage, though their impact is more difficult to quantify \citep{Kwan2001, AKCAOGLU2017376, Karaguler2018,bisschop-2002}. A robust statistical evaluation of these parameters is challenging due to the scarcity of real-life data and lengthy experimental processes. In this regard, numerical simulations offer a remedy, and several techniques of multiscale modelling of drying shrinkage were developed over the past years \citep{PICHLER200734, HAVLASEK2016, LIU201788, ABDELLATEF2019629}. However, the full-field evaluation of shrinkage and related mortar damage remains computationally costly.

To combat the aforementioned issue, we employ a deep-learning-based surrogate which can evaluate time-dependent full-field damage on microstructure samples, as well as predict the related homogenised properties, such as observed shrinkage and residual stiffness. The trained model can then be used to efficiently compute statistics on very large datasets to establish the relationship between the geometry of the microstructure and the related homogenised properties, opening a way to minimise internal damage and increase durability by changing the composition of the aggregate.

Since their introduction, deep learning models with convolutional layers, specifically U-Nets \citep{Ronneberger2015}, demonstrated their effectiveness in image-to-image tasks in a wide variety of applications, especially medical segmentation \citep{10643318}. U-Nets have become ubiquitous for concrete segmentation tasks as well, from labelling phases \citep{ZHOU2019144, Li2020, BANGARU2022104602, CHENIOUR2024134392, WERNER2025100416} to identifying cracks \citep{LIU2019129, Qiao2021, Rao2022, YU2022100436, 10562272, 10535162, Zhang2025} and other types of damage \citep{app13042398}. Additionally, the U-Net-based architectures have proven to be effective for a variety of different tasks beyond segmentation, such as load transfer path analysis \citep{ZHAO2020101184}, groundwater flow prediction \citep{TACCARI2022104169}, fluid flows prediction \citep{10066980}, or linear elastic stress estimation \citep{LANGCASTER2024104948}. Recently, a deep-learning method utilising U-Net architecture was proposed to predict full-field, time-dependent damage at the mesoscale \citep{NAJAFIKOOPAS2025110675}. In this paper, we utilise a similar framework, but instead of focusing on fracture and crack dynamics under tensile loading, we concentrate on shrinkage-induced damage.

The article is structured as follows. First, we introduce the physical setting and the numerical model for predicting the full-field damage in concrete and the related homogenised properties (Section \ref{sec:model}), and discuss the data generation process (Section \ref{sec:generation}). The employed deep neural network (DNN) architectures are described in Section \ref{sec:architecture}. Performance of the proposed models is examined in Section \ref{sec:experiments}. The statistical analysis of the relationship between aggregate characteristics and effective mixture properties is discussed in Section \ref{sec:big_data}. The final Section~\ref{sec:conclusion} summarizes the main takeaways of the paper and provides resources for implementation, such as the code and datasets.

\section{Computational Framework}\label{sec:model}
This section presents the computational framework underlying our deep learning approach, comprising the finite element model used to generate training data and the procedure for creating large synthetic datasets. We first describe the constitutive models used to simulate shrinkage-induced damage in concrete microstructures (Section \ref{sec:general_model}) and introduce modeling assumptions and material parameters (Section \ref{sec:modeling_assumptions}).
Next, two representative scenarios are described: uniform shrinkage representative of interior concrete regions (Section \ref{sec:uniform-simul}), and non-uniform shrinkage modeling the behavior of drying concrete surfaces (Section \ref{sec:nonuniform-simul}). Finally, we outline the level-set-based methodology used to generate the large collection of synthetic microstructures required for neural network training (\ref{sec:generation}).

\subsection{Microstructural constitutive models}\label{sec:general_model}
Linear elastic material described by its Young's modulus $E_a$ and Poisson's ratio $\nu_a$ is used to specify the behaviour of aggregates. To achieve computational efficiency and robustness, an isotropic damage model is applied to capture the tensile cracking of the matrix and the interfacial transition zone (ITZ).

In this model, the nominal stress tensor is computed from the total strain tensor $\boldsymbol{\varepsilon}$ as
\begin{equation}
    \boldsymbol{\sigma} = \left( 1- \omega \right) \tilde{\boldsymbol{\sigma}} = \left( 1- \omega \right) : \boldsymbol{D_e} \boldsymbol{\varepsilon}, 
\end{equation}
where $\omega$ is a scalar damage variable, $\tilde{\boldsymbol{\sigma}}$ is the effective stress tensor, and $\boldsymbol{D_e}$ is the stiffness tensor of an elastic material, depending on the elastic constants $E$ and $\nu$.

To model the resulting damage of the matrix and ITZ, an exponential softening law is used, described by two additional parameters: the tensile strength $f_t$ and the fracture energy $G_f$.
To provide objective results of the finite element simulation, the model response is regularised by the classic crack-band approach~\citep{Bazant-Oh}. Therefore, the expression for the damage variable $\omega$ is implicitly given by the following nonlinear equation
\begin{equation}
( 1- \omega ) E \kappa = f_t \exp \left( -\frac{h \omega \kappa f_t}{G_f} \right),
\end{equation}
where $h$ corresponds to the projection of the finite element in the direction of the maximum principal stress at crack initiation (assumed to be equal to the crack band width), and $\kappa$ is the maximum value of the equivalent strain $\tilde{\varepsilon}$ reached so far during the simulation. 

The Rankine failure criterion, smoothed in the region of multi-axial tension, is used to evaluate the equivalent strain
\begin{equation}
\tilde{\varepsilon}  = \frac{1}{E}\sqrt{\sum_{I = 1}^3 \langle \sigma_I \rangle^2},
\end{equation}
where $\langle\cdot\rangle$ denotes the positive part of the argument and $\sigma_I$ are the principal values of the effective stress tensor.

\subsection{Modeling assumptions and material properties}\label{sec:modeling_assumptions}
Consistent with established mesoscale studies of concrete \citep{GRASSL201085, MARUYAMA201682, ZHOU2019116785, Idiart2011}, a three-phase representation is adopted, consisting of coarse aggregates, a mortar matrix and the Interfacial Transition Zone (ITZ). The finite element analysis is performed under the assumption of plane stress. For all simulations, a $32 \times 32$~mm RVE is discretised using a regular mesh of $100 \times 100$ quadrilateral elements. The ITZ is represented by a single layer of elements at the aggregate-matrix interface. Because the finite element size of 0.32~mm is significantly greater than the ITZ thicknesses typically reported in the literature~\citep{OLLIVIER199530,Scrivener2004}, the Young modulus and the shrinkage rate are set equal to those of the mortar. Similarly to studies dealing with the mesoscopic analysis of concrete \citep{GRASSL201085,Idiart-2012} and experiments \citep{Hsu}, a 25\% reduction in tensile strength and fracture energy was defined. With the present discretization, a higher reduction in stiffness would have resulted in excessive compliance and macroscopic shrinkage, as demonstrated in \citep{SONG2024138112}.

The material parameters are kept constant throughout all simulations to isolate the effect of aggregate topology. A mesoscopic model, using Young's modulus for aggregates, $E_a = 50$~GPa, and equal stiffness for mortar and ITZ, $E_m = E_{itz} = 25$~GPa, yields an overall stiffness of around 30~GPa, which is a typical value for material~\citep{neville}. A Poisson’s ratio of $\nu = 0.2$ is assumed for all phases. For mortar, typical values of uniaxial tensile strength $f_t = 4$~MPa and fracture energy $G_f = 100$~N/m are used.

The complete list of material parameters used to describe the individual constituents of the model is given in Table~\ref{tab:param_structural}.

\begin{table} [!htb]
    \caption{Summary of material parameters in the structural analysis.}
    \label{tab:param_structural}
   \centering
   \begin{tabular}{ l  c  c  c}
   \toprule
   &  Parameter & Unit & Value \\
   \cmidrule{2-4}
   \multirow[t]{4}{*}[-1em]{Matrix}
   & $E_m$ & GPa & 25 \\
   & $\nu_m$ & -- & 0.2 \\
   & $f_{t,m}$ & MPa & 4 \\
   & $G_{f,m}$ & N/m & 100 \\
   \cmidrule{2-4}
   \multirow[t]{3}{*}[-0.5em]{Coarse aggregates}
   & $E_a$ & GPa & 50\\
   & $\nu_a$ & -- & 0.2 \\
   \cmidrule{2-4}
   \multirow[t]{3}{*}[-0.5em]{ITZ}
   & $E_{itz}$ & GPa & 25 \\
   & $\nu_{itz}$ & -- & 0.2 \\
   & $f_{t,itz}$ & MPa & 3 \\
   & $G_{f,itz}$ & N/m & 75 \\
  \bottomrule
   \end{tabular}
\end{table}

Given that the primary objective of this study is to isolate the influence of aggregate geometry and spatial distribution, other factors---including variability in material properties and time-dependent phenomena such as creep or specific drying rates---are disregarded. Nevertheless, this approach still allows for a direct comparison of the mechanical responses induced by different coarse aggregate topologies \citep{MARUYAMA201682, GRASSL201085}.

To maximise computational efficiency, the moisture transport problem is omitted. Instead, the evolution of shrinkage strain is explicitly prescribed for the mortar and ITZ phases, which are considered to be the only humidity-dependent constituents. The stiff coarse aggregates are treated as volumetrically stable, resulting in an incompatible stress field that induces tensile cracking in the surrounding phases. Since constitutive models are time-independent, the concept of pseudo-time $t$ is used to parameterize loading by increasing shrinkage eigenstrain.

All numerical simulations were run using non-linear statics in an open-source finite element solver OOFEM~\citep{OOFEM,OOFEMPatzak}. 

\subsection{Shrinkage scenarios}\label{sec:scenarios}
We consider two fundamental scenarios inspired by recent experimental studies \citep{HAVLASEK_creep_2021,HAVLASEK_creep_2025}, which focused on drying shrinkage-induced deformation of 100-mm-deep unreinforced concrete beams with different drying configurations. Before drying, the specimens were moist-cured for one month to achieve a uniform distribution of the material properties and an advanced degree of hydration, consistent with the use of Portland cement. Furthermore, autogenous shrinkage is disregarded, as it was effectively compensated for by the moist-curing conditions and would have largely occurred before the commencement of the drying experiments. For these reasons, it is assumed that all material properties remain constant throughout the simulation.

After four years of drying, the concrete beams exhibited significant stiffness reductions compared to geometrically identical sealed reference specimens: 12.5\% (top surface drying), 17.8\% (bottom surface drying), and 24.4\% (top and bottom surfaces drying). In \citep{HAVLASEK_creep_2024} this reduction was primarily attributed to tensile cracking caused by internally restrained drying shrinkage. 

In the first scenario, the Representative Volume Element (RVE) is located in the centre of the concrete beam and is subjected to gradually increasing shrinkage. Within the RVE, the shrinkage of the cement matrix is assumed to be spatially uniform, while the aggregates—which impose internal restraint—remain volume stable.

Conversely, in the second scenario, the RVE is considered at the outer edge of the beam, and thus exposed to a non-uniform drying, resulting in differential shrinkage across the mesoscale geometry. The distinctive details of each setting are elaborated in Sections~\ref{sec:uniform-simul} and \ref{sec:nonuniform-simul}. 

\subsubsection{Scenario 1. Uniform shrinkage}
\label{sec:uniform-simul}
\begin{figure}[h]
    \centering
    \includegraphics[width=0.9\textwidth]{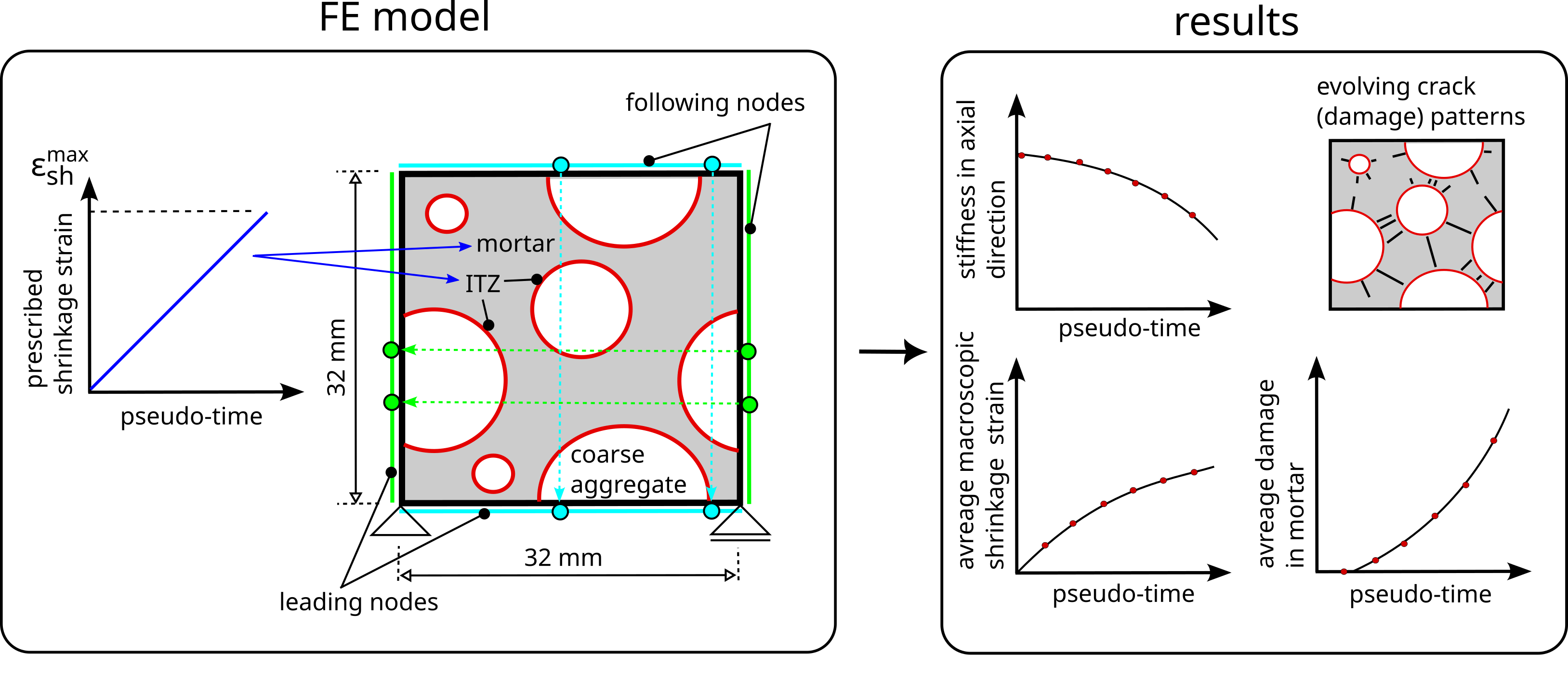}
    \caption{Schematic representation of the modelling approach for the simulation of a concrete mesostructure subjected to uniform shrinkage of mortar restrained by elastic aggregates (Uniform shrinkage scenario, Section~\ref{sec:uniform-simul}).}
    \label{fig:uniform_shrinkage}
\end{figure}

The first computational scenario, schematically described in Fig.~\ref{fig:uniform_shrinkage}, considers an RVE located in the central part of a drying concrete section. In this region, the highly nonlinear moisture diffusivity of concrete results in a nearly uniform humidity distribution that decreases over time. Consequently, the RVE is subjected to a spatially uniform shrinkage of the cement matrix. Although primarily representing the interior of a drying concrete member, this scenario is also applicable to modelling autogenous shrinkage, which is generally considered to be also spatially uniform. In this setting, the restraint arises only from the strain incompatibility between the mortar and the aggregates. This contrasts with the second scenario, which includes cross-sectional restraint. 

The model is externally statically determinate and the leader-follower concept is used to impose periodic boundary conditions on the lateral surfaces, as marked by different colours in Figure~\ref{fig:uniform_shrinkage}. The model is loaded by incrementing the negative isotropic shrinkage eigenstrain in the mortar and ITZ by $10~\mu\varepsilon$ per pseudo-time step, $t$, until reaching an ultimate value of $\varepsilon_{sh}^{\max} = 1000~\mu\varepsilon$~\citep{Maruyama2014} after 100 steps.

The quantities of interest are: the evolution of the macroscopic overall stiffness $k_u(t)$, (observed) macroscopic shrinkage $\varepsilon_{o,sh}(t)$, average damage $\Omega(t)$ (per volume of shrinking constituent), and the evolving crack pattern $\omega(\bm{x},t)$ as a function of an increasing shrinkage strain in pseudo-time $t$. Throughout the simulation, these quantities of interest are evaluated eleven times: at the beginning of the simulation to determine the initial state, and then every ten computational steps, i.e. after the shrinkage increment of 100~$\mu \varepsilon$. 

\subsubsection{Scenario 2. Non-uniform shrinkage}
\label{sec:nonuniform-simul}
\begin{figure}[h]
    \centering
    \includegraphics[width=0.9\textwidth]{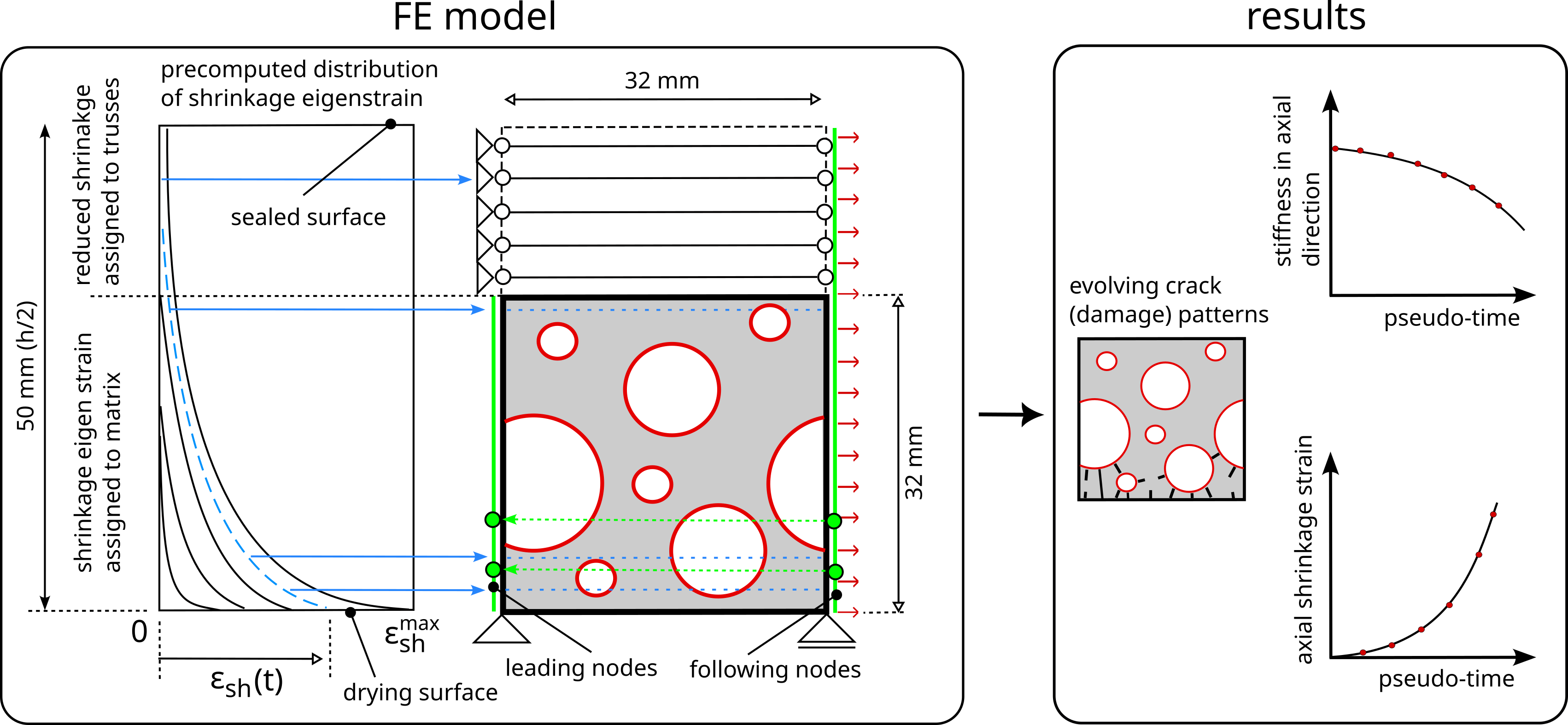}
    \caption{Schematic representation of the modelling approach for the simulation of a representative section of an internally restrained concrete beam subjected to nonuniform shrinkage of mortar (Nonuniform shrinkage scenario, Section~\ref{sec:nonuniform-simul}).}
    \label{fig:nonuniform_shrinkage}
\end{figure}
The second scenario, illustrated in Fig.~\ref{fig:nonuniform_shrinkage}, represents a 32-mm-long section of a 100~mm deep concrete beam that is drying symmetrically from its top and bottom surfaces, which allowed modeling only a symmetric half of the cross-section. The beam is not externally loaded and its axial deformation is unconstrained. 

This configuration introduces an additional source of restraint beyond the incompatibility between elastic aggregates and shrinking mortar. It arises from the nonuniform distribution of relative humidity, which results in a nonuniform profile of shrinkage eigenstrains. The strains are kinematically constrained by the assumption of cross-sectional planarity, leading to a self-equilibrated stress field with the exposed drying surface in tension and the core in compression. The magnitude of the tensile stresses increases with the degree of internal restraint and thus with the depth of the beam. The maximum principal stresses are initially oriented parallel to the drying surface, causing the first cracks to form perpendicularly to the surface. As drying progresses, cracks propagate between adjacent aggregates; previous studies suggest that crack widths increase when aggregates are larger and more closely spaced \citep{NEZERKA201749, NEZERKA2020118673}.

The corresponding FE model is composed of two parts: i) a regular $100 \times 100$ finite element mesh representing the mesostructure at the drying surface, with periodic boundary conditions imposed in the axial direction, and ii) truss elements representing the core. These truss elements mimic the structural stiffness of the beam core at a reduced computational cost, with material properties and shrinkage strains scaled to match the average response of the mesostructure.

\begin{figure}[h]
    \centering
    \includegraphics[width=0.5\textwidth]{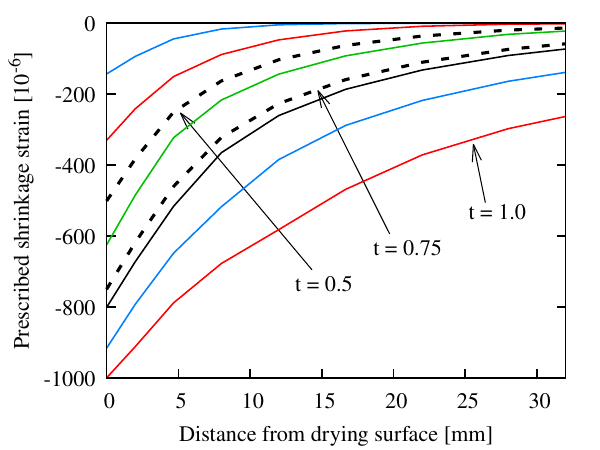}
    \caption{Shrinkage profiles prescribed to the mortar phase within the concrete RVE for the non-uniform shrinkage scenario (Section~\ref{sec:nonuniform-simul}). Solid colored lines represent scaled reference profiles exported from a hygro-mechanical FEM simulation \citep{HAVLASEK_creep_2025}, while the black dashed lines illustrate the linear interpolation performed for two selected values of pseudo-time.}
    \label{fig:shrinkage_profiles}
\end{figure}

The shrinkage profiles used for loading represent the first month of drying, a period characterised by the most severe eigenstrain gradients and by the maximum deflection of the beams subjected to non-symmetric drying~\citep{HAVLASEK_creep_2021,HAVLASEK_creep_2025}. 
Although moisture transport was originally simulated~\cite{HAVLASEK_creep_2025} using Kuenzel’s model \citep{Kunzel} and the modified Microprestress-Solidification model \citep{HavlasekPhD} was adopted for the subsequent structural sub-problem, in this work the shrinkage profiles are explicitly prescribed and are not influenced by the mesostructure. The shrinkage profiles shown in Fig.~\ref{fig:shrinkage_profiles} were scaled by a factor of 1.88, such that the maximum shrinkage on the surface at the end of the simulation, i.e.~$t=1$, is equal to $\varepsilon_{sh}^{max} = 1000~\mu \varepsilon$, consistent with Scenario~1. During the simulation, the surface shrinkage increases linearly, while the internal distribution is determined via linear interpolation between the pre-calculated profiles, as shown for $t=0.5$ and $t=0.75$ by dashed black lines in Fig.~\ref{fig:shrinkage_profiles}. (When these shrinkage profiles are prescribed in simulations of non-symmetric drying with modified geometries, depths of 100, 150, and 200~mm, and boundary conditions permitting free rotation of the ends, the resulting curvature exhibits good agreement with the values experimentally measured after 1 month of drying.)

For training of surrogate models, we record the following quantities of interest: crack pattern, incremental stiffness $k_n(t)$ in the axial direction, and the resulting axial shrinkage strain. All quantities are exported for the initial state and then after increments of $100 \mu \varepsilon$.

\subsection{Data Generation}
\label{sec:generation}

To overcome the scarcity of real-world microstructural datasets and to enhance the predictive capabilities of the developed surrogate model, we generated a large collection of synthetic microstructures designed to reflect the variability in coarse aggregate geometry. This approach enabled training on a sufficiently diverse dataset, allowing the model to generalise across a wide range of mesostructural configurations and, ultimately, to evaluate the influence of aggregate topology and distribution on shrinkage-induced damage.

The synthetic microstructures were generated using an in-house tool~\citep{DOSKAR2020} based on the level-set method proposed by~\cite{Sonon2012AUL}. This method extends the traditional random sequential adsorption (RSA) approach to generating particulate media, e.g.,~\citep{KUBALA2022102692, Lebovka2023}. Unlike conventional RSA (also known as the dart-throwing algorithm), where candidate particles are randomly placed and subsequently rejected in case of overlap, the level-set-based implementation avoids overlap by construction. This is achieved by maintaining a signed distance field on a regular grid, which encodes the Euclidean distance to the nearest particle surface. Given the circumscribed radius of a new particle, candidate positions are sampled exclusively from regions where the distance exceeds this radius, guaranteeing non-overlapping placement a priori, rather than through ex-post checks. This method is described in more detail in \citep{DOSKAR2020}.

In addition to improved computational efficiency, especially at higher volume fractions, the level-set-based scheme enables finer control over particle dispersion. For instance, minimal and maximal distances to the nearest and next-nearest neighbours can be imposed, thereby tailoring the microstructure statistics more precisely to resemble those found in real concrete. Finally, the level-set approach enables generating complex geometries by switching to implicit microstructural description and computing linear combinations of distances to first-, second-, and third-nearest particles~\citep{Sonon2012AUL, DOSKAR2020}. %For the purpose of this work, we utilized the implicit geometry to smoothen particle vertices; see Section~\ref{sec:big_data}.

For each of the two scenarios, \numprint{15,000} unique geometries representing (semi-)periodic RVE using the level-set method implementation were generated. In both scenarios, the RVE is of size $32 \times 32$ mm and discretised to a $100 \times 100$ pixel format. The microstructural geometry consisted of three phases: mortar, modelled as a homogeneous mixture of cement paste and small aggregate particles; coarse aggregates (4-16 mm in diameter); and the ITZ (1 pixel wide). To cover a large variety of geometries, aggregate particles were modelled as polygons with four to eight vertices (chosen randomly from a uniform distribution). Each particle is first generated as a regular polygon inscribed into a circle of a given radius $r$. The angles between vertices are then randomised, as well as the distance of each vertex from its respective circle centre, ranging from $0.75r$ to $1r$. The largest particles (16 mm in diameter) are inserted first, followed by the medium-sized (8 mm) and, finally, the smallest particles (4 mm).

Since the RVE in the first scenario is assumed to be periodic and the damage and related homogenised properties do not depend on the sample orientation, the dataset was artificially enlarged by applying data augmentation. Several operations are randomly applied to the training dataset of \numprint{12,000} samples: orthogonal rotation, horizontal and vertical mirroring, and vertical and horizontal pixel shift. In the second scenario, with the semi-periodic RVE, the training data could be augmented only with random horizontal mirroring and horizontal pixel shift. 

\begin{figure}[h]
    \centering
    \includegraphics[width=\textwidth]{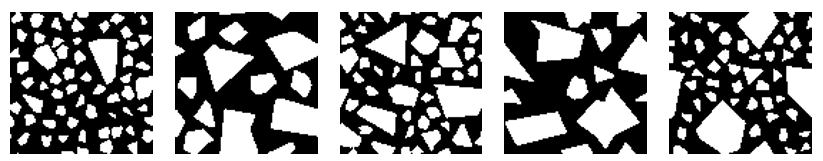}
    \caption{Examples of the generated microstructures with the level-set method.}
    \label{fig:microstructures}
\end{figure}

%The emphasis is given primarily on the effect of the coarse aggregate geometry and distribution; other aspects, such as variability in material properties or the time-dependent phenomena, are disregarded. Moreover, to maximise computational efficiency, the evolution of shrinkage strain is prescribed explicitly to the mortar and ITZ phases, which are considered to be the only humidity-dependent constituents. Because the viscoelastic nature of the problem is ignored, the effect of drying rate on cracking cannot be assessed. Nevertheless, this approach still enables one to directly compare responses obtained with different topologies of coarse aggregates. All numerical simulations were run using non-linear statics in an open-source finite element solver OOFEM~\citep{OOFEM,OOFEMPatzak}. 

\section{Network Architecture}\label{sec:architecture}

\begin{figure}[h]
    \centering
    \includegraphics[width=\textwidth]{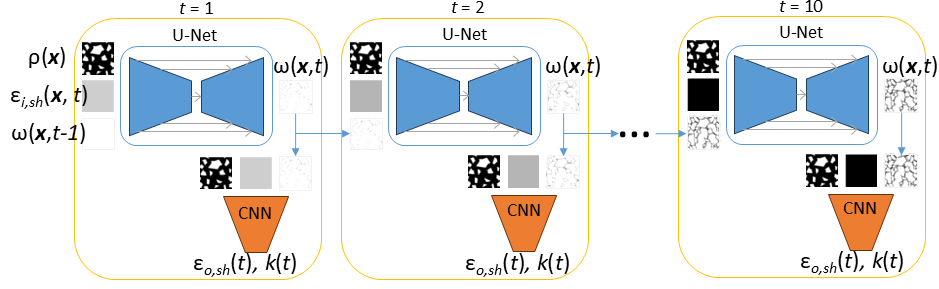}
    \caption{Auto-regression scheme. Two networks, a U-Net (depicted in blue) and a CNN (depicted in orange) are used to predict time-dependent full-field damage $\omega(\bm{x},t)$, macroscopic shrinkage $\varepsilon_{o,sh}(t)$, and effective residual stiffness $k(t)$. The output at the pseudo-time step $t$ is used as an input at the pseudo-time step $t+1$.}
    \label{fig:autoregression_scheme}
\end{figure}

Our approach utilises two network architectures to predict the time series of the full-field damage of a unit cell and the related homogenised properties, see Fig.~\ref{fig:autoregression_scheme}. The first network, in the form of an auto-regressive U-Net, predicts the scalar damage $\omega(\bm{x},t)$ field at the pseudo-time step $t$ from a given geometry field $\rho(\bm{x}) \in \{0,0.5,1\}$, imposed shrinkage field $\varepsilon_{i,sh}(\bm{x},t)$ at the pseudo-time step $t \in \{1, \dots, 10\}$, and damage field $\omega(\bm{x},t-1)$ of the previous step $t-1$. The predicted damage at step $t$ is then used as an input at the next step $t+1$. The second network, a standard convolutional neural network (CNN) uses the damage field $\omega(\bm{x},t)$ predicted by the U-Net along with the geometry and imposed shrinkage field $\varepsilon_{i,sh}(\bm{x},t)$ to predict the observed macroscopic shrinkage $\varepsilon_{o,sh}(t)$ and the effective residual stiffness $k(t)$. Both adopted architectures are detailed below. 

\subsection{Auto-regressive U-Net}\label{sec:unet}

\begin{figure}[h]
    \centering
    \includegraphics[width=\textwidth]{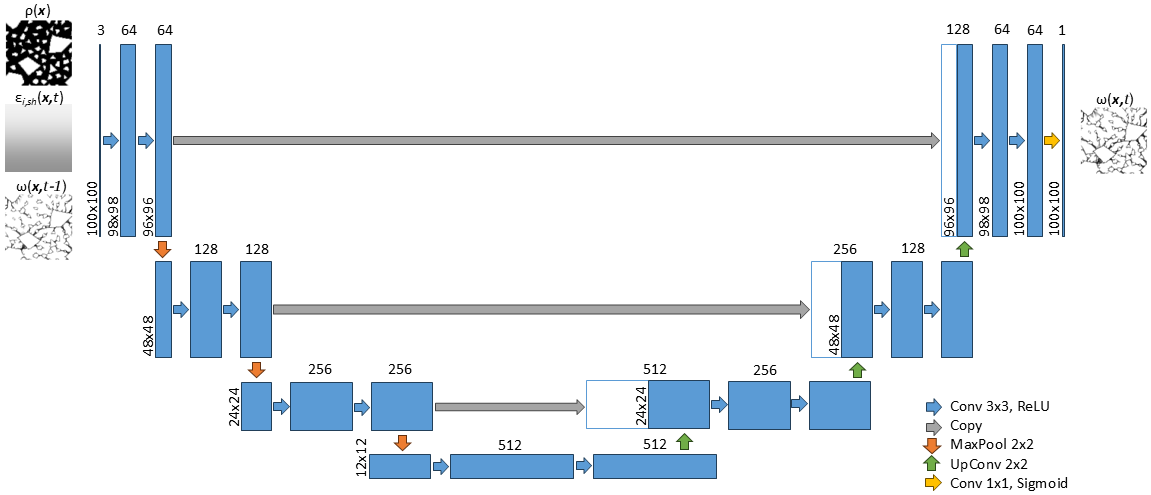}
    \caption{Architecture of the adopted auto-regressive U-net with 7,697,345 parameters. The network takes a multi-channel image, where the first channel represents the geometry $\rho(\bm{x})$ (1 for mortar, 0.5 for interface, 0 for aggregate); the second channel contains the normalized imposed shrinkage $-\varepsilon_{i,sh}(\bm{x},t)/0.001$ (continuous values between [0, 1]); and the third channel represents the damage field $\omega(t-1)$ (continuous values between [0, 1]) of the previous time step. Blue rectangles represent multi-channel feature maps. The number of channels is denoted on top of the rectangle. The size of the feature map is provided at the lower left edge of the rectangles. Copied feature maps are represented by white rectangles. The colored arrows denote different operations, described in the legend.}
    \label{fig:autounet_scheme}
\end{figure}

The U-Net model is a type of CNN architecture, named for its distinctive U-shape \citep{Ronneberger2015}, which is achieved through a symmetric arrangement of layers. The architecture can be divided into two main parts, cf. Fig.~\ref{fig:autounet_scheme}:

\begin{itemize}
    \item Downsampling path, consisting of a sequence of convolutional and pooling layers. The convolutions are used to capture the context in the image, which helps in understanding features at various scales. The pooling operations reduce the spatial dimensions of the feature maps, increasing the field of receptive view and enabling the network to capture more global features of the input image.

    \item Upsampling path, consisting of a series of up-convolutions and concatenations followed by regular convolutional layers. The up-convolutions effectively increase the spatial dimensions of the feature maps. A key feature of U-Net is the concatenation of these upsampled outputs with the feature maps from the corresponding layers in the downsampling path (skip connections). This process helps the network to localise and use precise spatial information from earlier steps in the network.
\end{itemize}

Here, we adapt the standard U-Net architecture to iteratively process the time-dependent information from the concrete shrinkage simulation. The general scheme of the network is presented in Fig.~\ref{fig:autounet_scheme}. The U-Net is fed a multi-channel image as input, similar to RGB images in the segmentation tasks. In this case, the channels represent different information about the microstructure and loading conditions. The first channel encodes the microstructure $\rho(\bm{x})$, where pixels belonging to the mortar are assigned the value 1, pixels of the aggregate are assigned the value 0, and the ITZ pixels are assigned the value 0.5. The second channel represents the imposed shrinkage load $\varepsilon_{i,sh}(\bm{x},t)$ in the step $t$ on the sample, which evolves in time and can be non-uniform (cf. Section~\ref{sec:nonuniform-simul}). For faster training, it is normalised by the maximum absolute value at the last step $\varepsilon_{i,sh}(\bm{x},t=10)$ to contain values between 0 and 1, so it is of similar order of magnitude to other channels. Finally, the third channel contains the accumulated damage $\omega(\bm{x},t-1)$ predicted in the previous step $t-1$, represented by continuous values between 0 and 1 (0 -- no damage, 1 -- full damage). Since there is no initial damage in step $t=1$, this channel starts with all values set to 0. 

The U-Net outputs an image of the accumulated damage field $\omega(\bm{x},t)$ of continuous values between 0 and 1. To ensure that $\omega(\bm{x},t+1)$ is larger than $\omega(\bm{x},t+1)$, the output is set to $\max\,[{\omega(\bm{x},t), \omega(\bm{x},t+1)}]$. This prediction is then used as damage input for the same network at time step $t+1$. The imposed shrinkage is also updated for the next time step. This process repeats sequentially until the data for all time steps is obtained.

Unlike recurrent U-Net architectures, e.g., \citep{Wang2019}, our model does not have internal memory. Since the reference model only requires the information from the previous time step to compute the next step, cf. Section~\ref{sec:model}, we reasoned that having hidden states would be unnecessary. As a result, the GPU memory requirements for training the network are low, and the weights can be updated at every time step, speeding up the training process.

The loss of step $t$ is computed as the average of pixel-wise mean-squared errors between the ground truth damage and the prediction:

\begin{equation}
    \mathcal{L}_{\omega}^t = \frac{1}{n_{\text{pix}}} \sum_{p=1}^{n_{\text{pix}}} \left( \omega_p^{\text{ref}}(\bm{x},t) - \omega_p^{\text{nn}}(\bm{x},t) \right)^2, \; t = 1,2,\dots,10; 
\end{equation}
where $\omega_p^{\text{ref}}(x,t)$ is the reference value at the pixel $p$ and time step $t$, and $\omega_p^{\text{nn}}(x,t)$ is the corresponding network prediction.

\subsection{Convolutional network}\label{sec:convnet}
\begin{figure}[h]
    \centering
    \includegraphics[width=\textwidth]{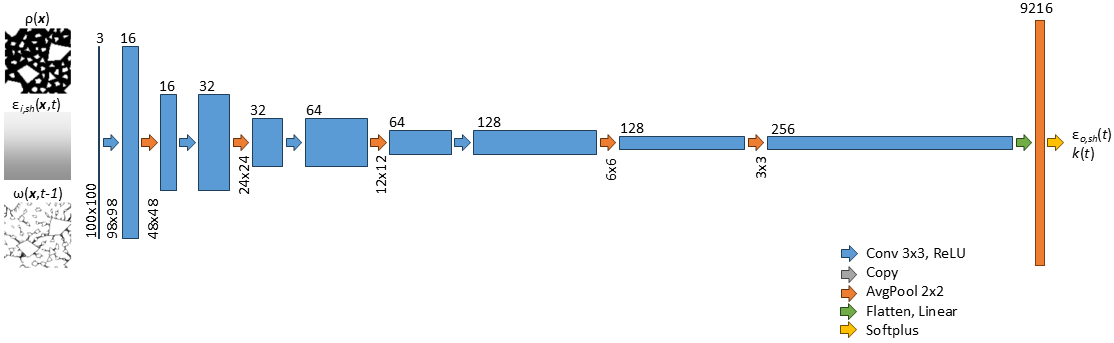}
    \caption{Architecture of the adopted convolutional network (CNN) with 411,042 parameters. Similar to the U-Net, the convolutional network takes a multi-channel image, representing the geometry $\rho(\bm{x})$, damage field $\omega(\bm{x},t)$, and normalised imposed shrinkage $\bar{\varepsilon}_{i,sh}(t)$. Blue rectangles represent multi-channel feature maps. The number of channels is denoted on top of the rectangle. The size of the feature map is provided at the lower left edge of the rectangles. The final feature maps are flattened (orange rectangle) and connected to the scalar outputs via a fully connected layer. The colored arrows denote different operations.}
    \label{fig:convnet_scheme}
\end{figure}

Since the U-Net part of the framework maps image-to-image microstructural data, we employ an additional convolutional network to extract the observed effective normalised quantities such as effective shrinkage $\bar{\varepsilon}_{o,sh}(t)$ and the residual stiffness $\bar{k}(t)$, see Fig.~\ref{fig:convnet_scheme}. The inputs at a step $t$ are almost identical to the U-Net, except that the current damage data for step $t$ is used in the third channel instead of the previous time step (either ground truth or predicted by the U-Net). The network gradually downsamples the feature maps with average pooling and then densely connects them to the two scalar outputs, which are normalised observed shrinkage and residual stiffness. The output is constrained by the softplus activation function to produce non-negative values. The loss at step $t$ is computed as

\begin{equation}
    \mathcal{L}^t = c \cdot \mathcal{L}_{\varepsilon}^t + \mathcal{L}_{k}^t = c \cdot|\varepsilon_{o,sh,n}^{\text{ref}}(t) - \varepsilon_{o,sh,n}^{\text{nn}}(t)| + |k_n^{\text{ref}}(t) - k_n^{\text{nn}}(t)|, \; t = 1,2,\dots10; 
\end{equation}
where $\bar{\varepsilon}_{o,sh}^{\text{ref}}(t)=-\varepsilon_{o,sh}^{\text{ref}}(t)/{\max}[ \varepsilon_{i,sh}(\bm{x},t=10)]$ and $\bar{k}^{\text{ref}}(t)=k^{\text{ref}}(t)/30\,\text{GPa}$ are the reference values for the normalised observed shrinkage and residual stiffness, respectively, and $\bar{\varepsilon}_{o,sh}^{\text{nn}}(t)$ and $\bar{k}^{\text{nn}}(t)$ are the network predictions, with the weighting constant $c = 1000$, found empirically.

\section{Experiments}\label{sec:experiments}

This section discusses the accuracy of the trained neural networks for both uniform and non-uniform shrinkage scenarios. In each case, every network was trained for 100 epochs with the ADAM algorithm with adaptive learning rate on NVIDIA GeForce RTX 4070 Ti with 12 Gb VRAM.

\subsection{Scenario 1. Uniform shrinkage}

As already discussed, the first scenario models a uniformly shrinking central part of a concrete beam. In order for the U-Net to be able to predict the periodic evolution of the damage profile, periodic padding was used in the downsampling path in both vertical and horizontal directions. 

\begin{figure}[H]
    \centering
    \includegraphics[width=\textwidth, trim={1.5cm 0 0 0},clip]{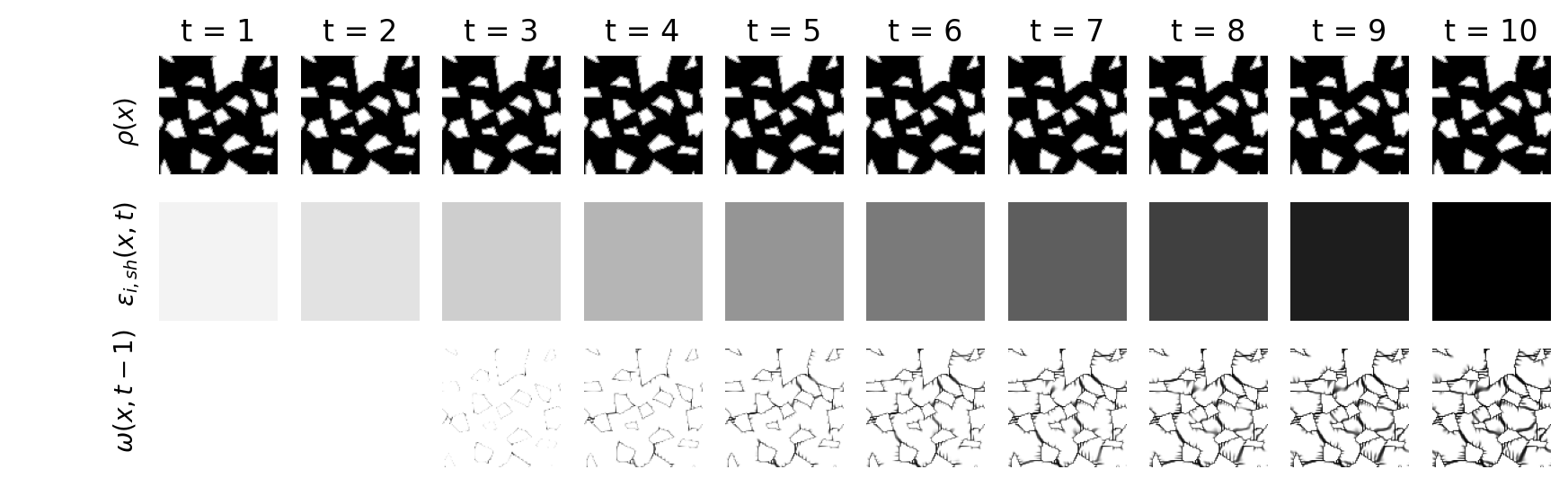}
    \caption{Example of the U-Net input channels for Scenario 1 for different time steps $t$. The first channel (top row) represents the geometry (1 for mortar, 0.5 for interface, 0 for aggregate), the second channel (middle row) contains the normalized imposed shrinkage $\bar{\varepsilon}_{i,sh}(t)  = \varepsilon_{i,sh}(t) / \varepsilon_{i,sh}(t=10)$(continuous values between [0, 1]), and the third channel (bottom row) represents the damage (continuous values between [0, 1]) in the previous step $t-1$. Since the imposed shrinkage is uniform in this scenario, it is not dependent on $\bm{x}$.}
    \label{fig:inputs_uniform}
\end{figure}

\begin{figure}[H]
    \centering
    \includegraphics[width=\textwidth]{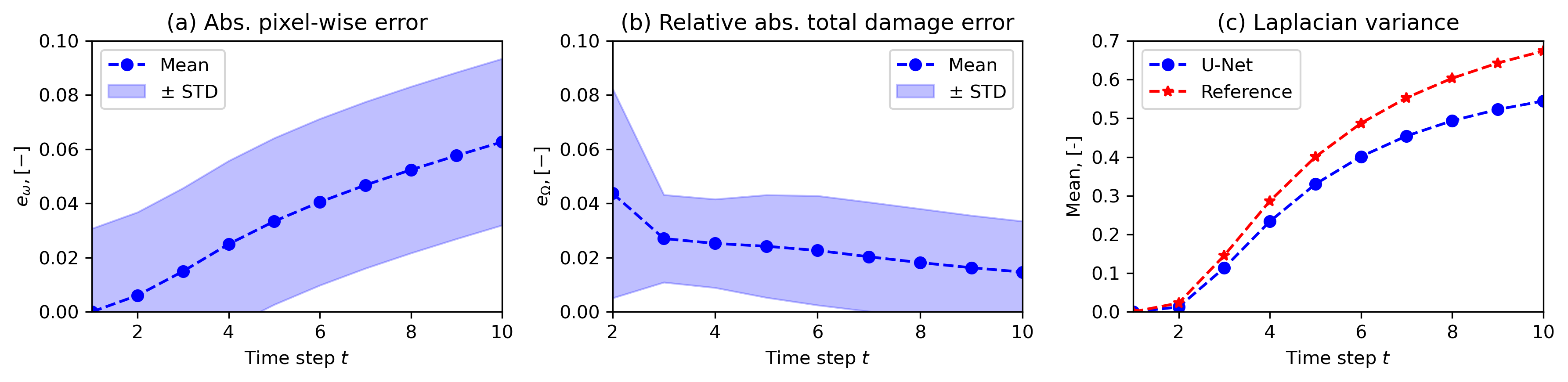}
    \caption{Mean (solid line) $\pm$ standard deviation (colored interval) of the prediction errors on the test dataset of Scenario 1. The absolute pixel-wise damage error $e_{\omega}$ (a) is averaged over mortar and interface pixels, excluding aggregate pixels. The total damage $\Omega(t)$ (b) represents the total ratio of damaged pixels in the RVE. The average Laplacian variance (c) indicates the sharpness of the damage image.}
    \label{fig:error_damage_uniform}
\end{figure}

The input channel examples for Scenario 1 are presented in Fig.~\ref{fig:inputs_uniform}. The exact damage data was used as the input to train the CNN, while the U-Net was trained auto-regressively from its own outputs. The average absolute pixel-wise error, shown in Fig.~\ref{fig:error_damage_uniform}, is defined as
\begin{equation}
    e_{\omega}(t) = \frac{1}{\sum_{p=1}^{n_\text{pix}} \mathbb{I}_{\rho(\bm{x}) \neq 0}} \sum_{p=1}^{n_\text{pix}} \mathbb{I}_{\rho(\bm{x})  \neq 0} \cdot |\omega_p^{\text{ref}}(\bm{x},t) - \omega_p^{\text{nn}}(\bm{x},t)|, \; t = 1, 2,\dots,10;
\end{equation}
where $\mathbb{I}_{\rho(\bm{x}) \neq 0}$ is the indicator function for the matrix and ITZ pixels. We average the error only over the matrix and ITZ pixels, since the damage does not propagate through the inclusions. Here we see that the error is the highest in the final step, around 6\% on the test dataset. Fig.~\ref{fig:error_damage_uniform} further shows that the U-Net can be reliably used to estimate the total RVE damage $\Omega(t) = \frac{1}{n_\text{pix}} \sum_{i=p}^{n_\text{pix}} \omega_p(\bm{x},t)$ as well, with the average relative error, defined as 
\begin{equation}
e_{\Omega}(t) = \frac{1}{\Omega^{\text{ref}}(t)} |\Omega^{\text{ref}}(t) - \Omega^{\text{nn}}(t)|,
\end{equation}
below 5\%, and decreasing with the time steps.

\begin{figure}[H]
    \centering
    \includegraphics[width=\textwidth, trim={3cm 0 0 0}, clip]{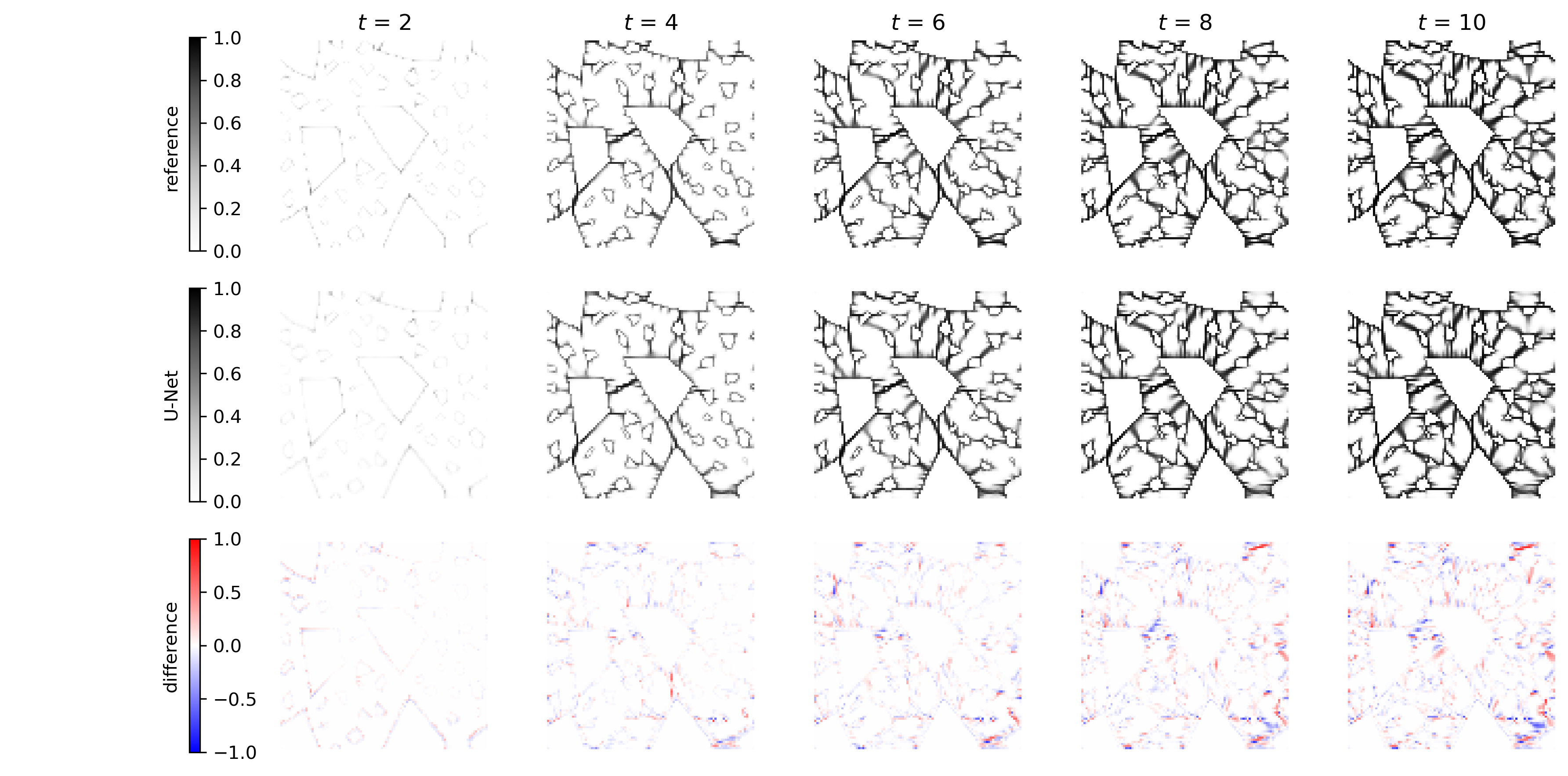}
    \caption{Example of the damage field prediction by the U-Net on a test geometry sample, Scenario 1. The first row shows the reference damage from the FEM simulation, the second row shows the U-Net prediction, and the third row shows the pixel-wise difference between the reference and the U-Net values.}
    \label{fig:damage_uniform}
\end{figure}

The U-Net has learned to produce images of damage fields that are visually very similar to the reference simulation, see the example in Fig.~\ref{fig:damage_uniform}. On closer inspection, the predicted localised damage profiles can be more dispersed or slightly shifted. This can be quantified as the difference between the average variance of the Laplacian \citep{7894491}, where a lower variance of the predicted images indicates a higher amount of blur (Fig.~\ref{fig:error_damage_uniform}c).

\begin{figure}[H]
    \centering
    \includegraphics[width=\textwidth]{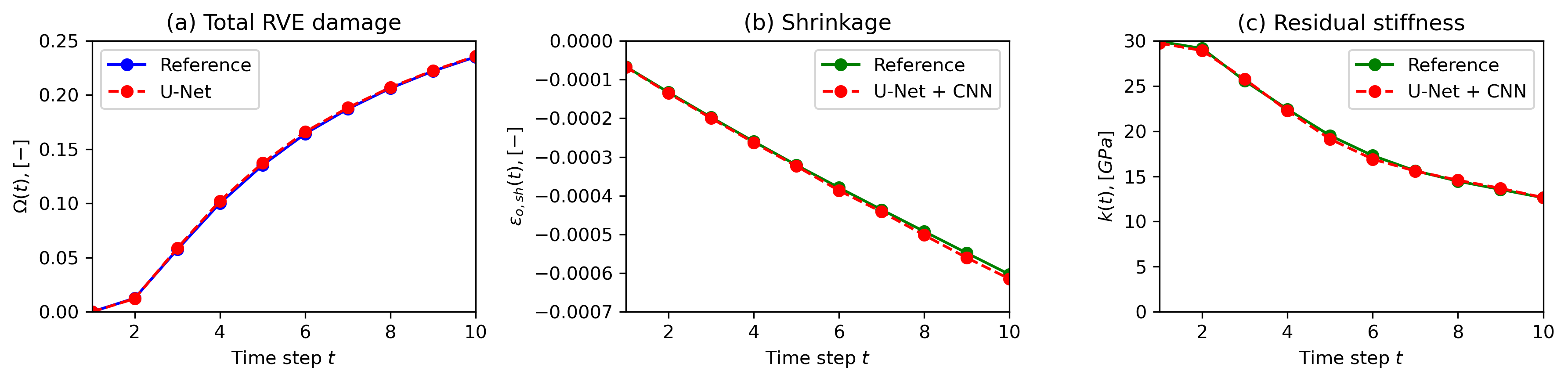}
    \caption{Prediction of the homogenised properties by the U-Net and CNN of the test geometry used in Fig.~\ref{fig:damage_uniform} as compared to the reference values, Scenario 1.}
    \label{fig:homog_uniform}
\end{figure}

\begin{figure}[H]
    \centering
    \includegraphics[width=0.66\textwidth]{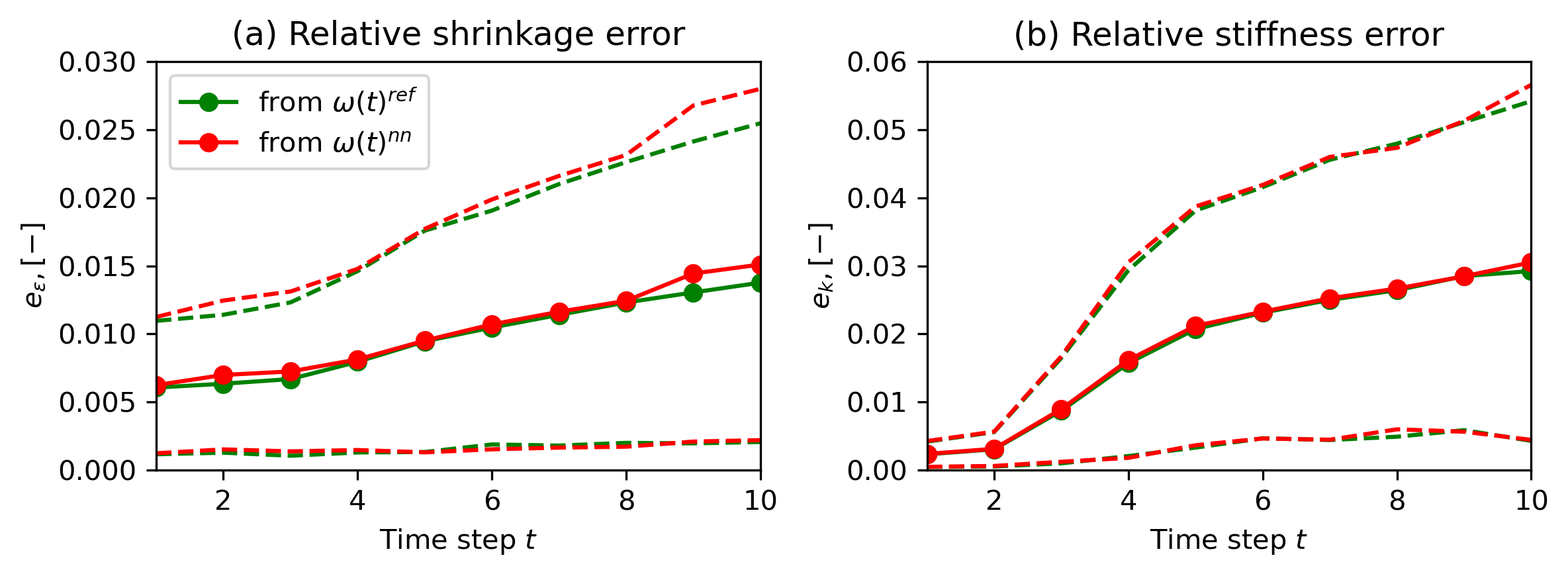}
    \caption{Mean (solid line) $\pm$ standard deviation (dashed line) of the relative prediction errors on the test dataset, Scenario 1. The errors for the shrinkage and residual stiffness prediction are calculated from the reference damage data (green) and from the damage predicted by the U-Net (red).}
    \label{fig:error_uniform}
\end{figure}

While the pixel-wise accuracy in predicting the damage field is above 90\%, it also a negligible effect on the CNN predictions of the observed shrinkage and residual stiffness. Figure~\ref{fig:homog_uniform} shows the comparison of the absolute predicted values for the geometry of Figure~\ref{fig:damage_uniform}. Figure~\ref{fig:error_uniform} then illustrates the average relative error for the observed shrinkage 
\begin{equation}
    e_{\varepsilon}(t)=\frac{|\varepsilon_{o,sh}^{\text{ref}}(t) - \varepsilon_{o,sh}^{\text{nn}}(t)|}{ |\varepsilon_{o,sh}^{\text{ref}}(t)|},
\end{equation}
and the average relative error for residual stiffness
\begin{equation}
e_{k}(t)=\frac{|k^{\text{ref}}(t) - k^{\text{nn}}(t)|}{k^{\text{ref}}(t)},
\end{equation}
across the entire test dataset. When the CNN utilises the U-Net for damage input, the relative average error of the homogenised properties at any given time step remains below 3\% for both shrinkage and residual stiffness.

\begin{figure}[H]
    \centering
    \includegraphics[width=0.75\textwidth]{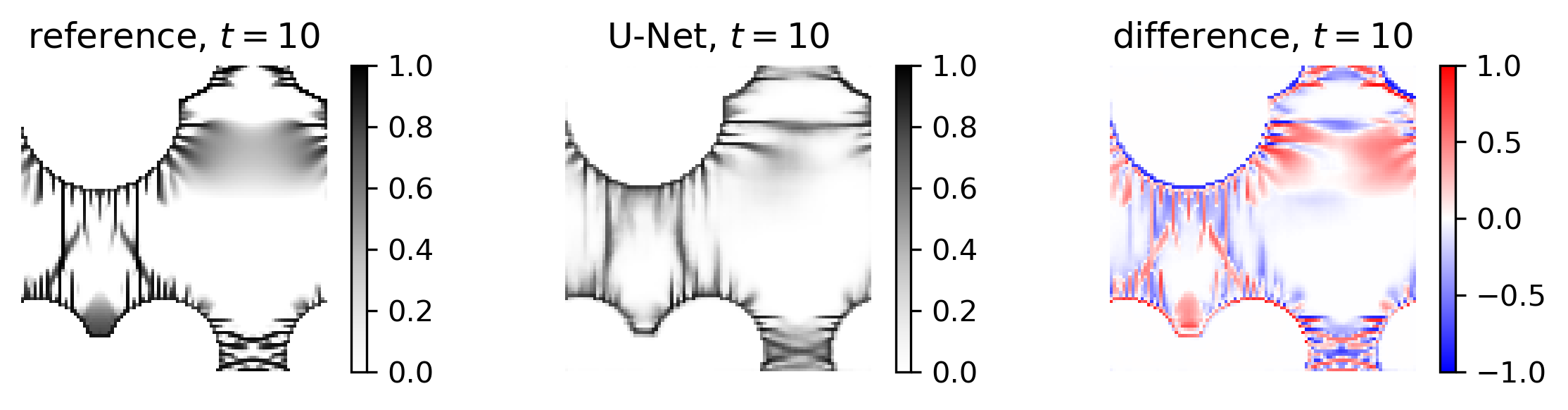}
    \caption{Example of the pixel-wise damage prediction by the U-Net on an arbitrary geometry of three overlapping circles.}
    \label{fig:damage_mickey}
\end{figure}

\begin{figure}[H]
    \centering
    \includegraphics[width=\textwidth]{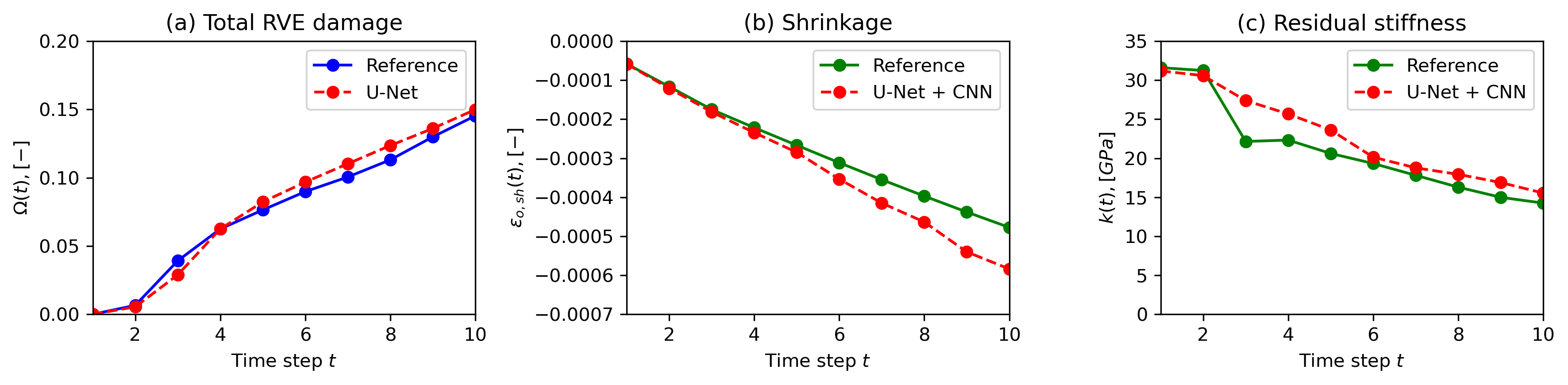}
    \caption{Prediction of the homogenised properties by the U-Net and CNN of the arbitrary geometry of three overlapping circles depicted in Fig.~\ref{fig:damage_mickey} compared to the reference values, Scenario 1.}
    \label{fig:homog_mickey}
\end{figure}

\begin{figure}[H]
    \centering
    \includegraphics[width=0.75\textwidth]{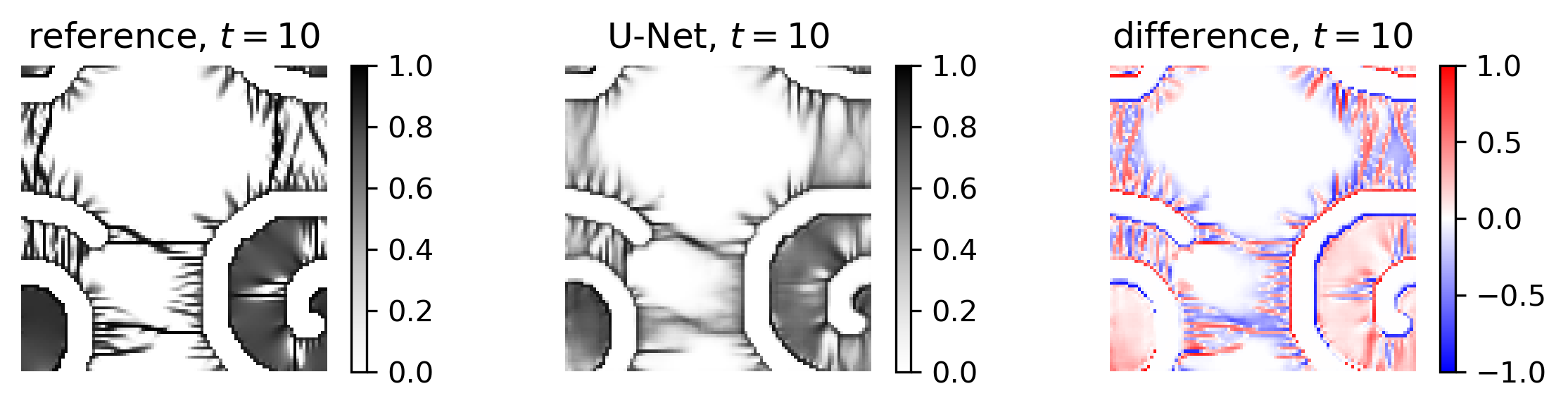}
    \caption{Example of the pixel-wise damage prediction by the U-Net on an arbitrary geometry of a hand-drawn spiral.}
    \label{fig:damage_spiral}
\end{figure}

\begin{figure}[H]
    \centering
    \includegraphics[width=\textwidth]{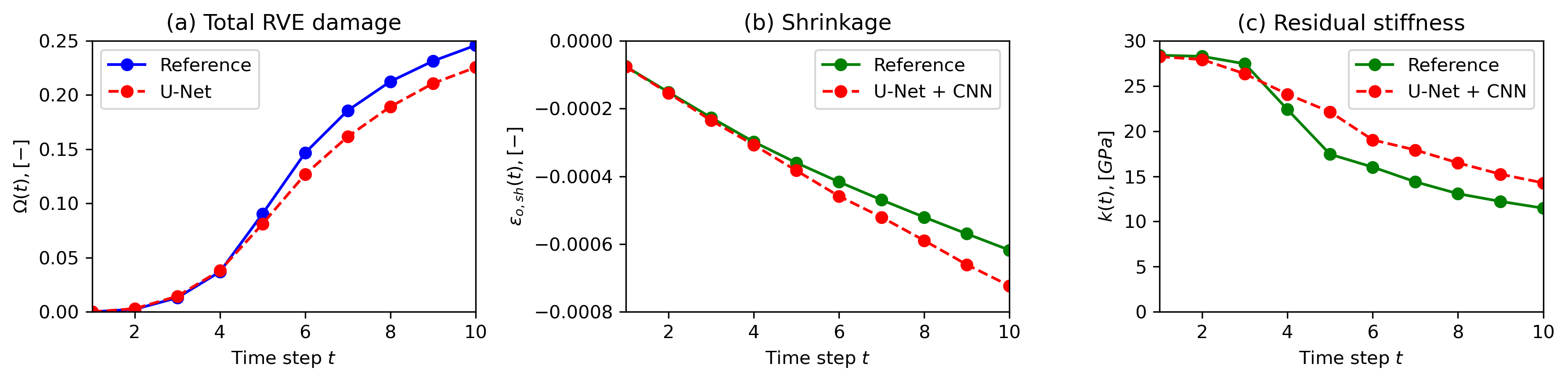}
    \caption{Prediction of the homogenised properties by the U-Net and CNN of the arbitrary geometry of a hand-drawn spiral in Fig.~\ref{fig:damage_spiral} compared to the reference values, Scenario 1.}
    \label{fig:homog_spiral}
\end{figure}

The trained U-Net and CNN architecture also show a relatively good capability to predict the responses of arbitrary geometries that are starkly different from those generated by the level-set method (Section~\ref{sec:generation}), which will be crucial for the data exploration study presented later in Section~\ref{sec:big_data}. We illustrate extrapolation capabilities of the networks on two artificial geometries, three overlapping circles (Fig.~\ref{fig:damage_mickey}) and a hand-drawn spiral (Fig.~\ref{fig:damage_spiral}). Both problems deal with a non-convex geometry with an unusual aggregate distribution. While the predictions are less accurate than what is typically observed in the test dataset (particularly the CNN component, while the U-Net struggles to keep longer damage paths sharp), the overall trends are still captured, with the relative error for damage prediction is under 10\%, the shrinkage error is below 20\%, and the error in residual stiffness is less than 25\% for all the time steps (Figs.~\ref{fig:homog_mickey}, ~\ref{fig:homog_spiral}). 

\subsection{Scenario 2. Non-uniform shrinkage}

\begin{figure}[H]
    \centering
    \includegraphics[width=\textwidth, trim={1.5cm 0 0 0},clip]{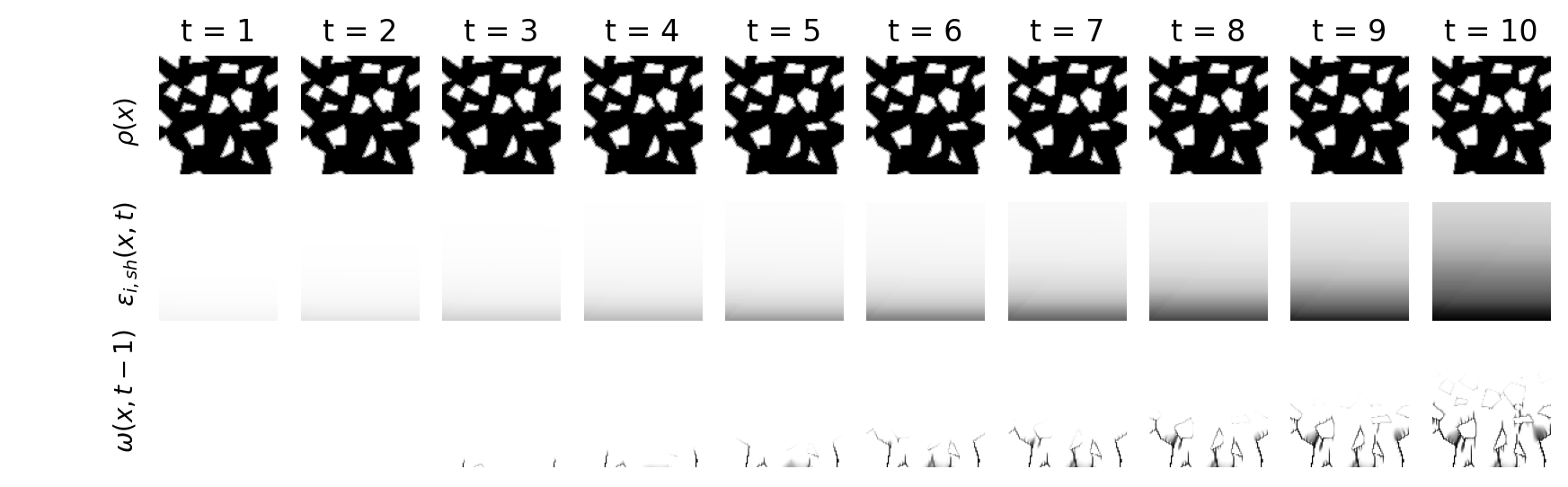}
    \caption{Example of the U-Net input channels for Scenario 2 for different time steps $t$. The first channel (top row) represents the geometry (1 for mortar, 0.5 for interface, 0 for aggregate), the second channel (middle row) contains the normalized imposed shrinkage $\varepsilon_{i,sh}(t) / \varepsilon_{i,sh}(10)$(continuous values between [0, 1]), and the third channel (bottom row) represents the damage (continuous values between [0, 1]) in the previous time step.}
    \label{fig:inputs_nonuniform}
\end{figure}

The same U-Net and CNN architectures, as depicted in Figs.~\ref{fig:autounet_scheme} and \ref{fig:convnet_scheme}, were used to estimate damage and the related properties for the dataset generated in Scenario 2, recall Section~\ref{sec:nonuniform-simul}, i.e., for non-uniform imposed shrinkage profiles. The main difference from Scenario 1 is the non-uniform imposed shrinkage that varies vertically throughout the RVE, see Fig.~\ref{fig:inputs_nonuniform}. In addition, the RVE is assumed to be only horizontally periodic, as its bottom edge represents the surface of the beam in contact with the air. Consequently, the periodic padding is applied only in the horizontal direction, and replication padding is used in the vertical direction instead.  

\begin{figure}[H]
    \centering
    \includegraphics[width=\textwidth]{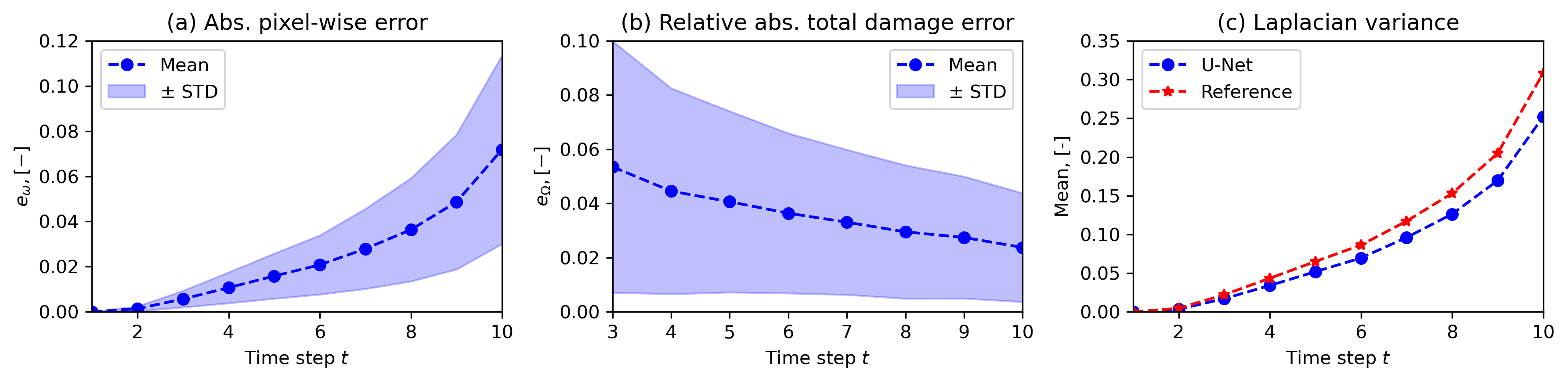}
    \caption{Mean (solid line) $\pm$ standard deviation (colored interval) of the prediction errors on the test dataset of Scenario 2. The absolute pixel-wise damage error $e_{\omega}$ (a) is averaged over mortar and interface pixels, excluding aggregate pixels. The total damage $\Omega(t)$ (b) represents the total ratio of damaged pixels in the RVE. The average Laplacian variance (c) indicates the sharpness of the damage image.}
    \label{fig:error_damage_nonuniform}
\end{figure}

\begin{figure}[H]
    \centering
    \includegraphics[width=\textwidth, trim={3cm 0 0 0}, clip]{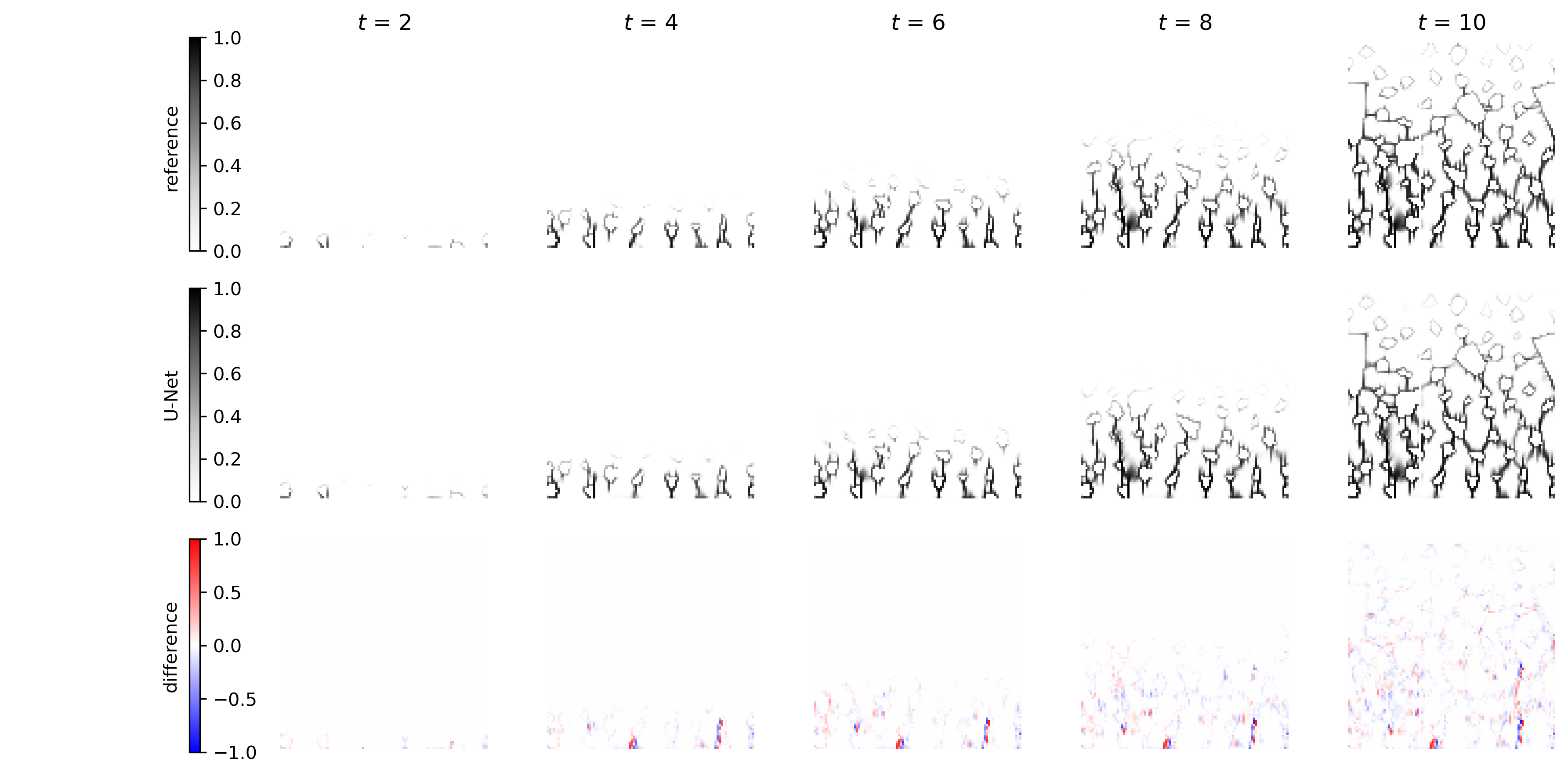}
    \caption{Example of the damage field prediction by the U-Net on a test geometry sample, Scenario 2. The first row shows the reference damage from the FEM simulation, the second row shows the U-Net prediction, and the third row shows the pixel-wise difference between the reference and the U-Net values.}
    \label{fig:damage_nonuniform}
\end{figure}

The networks were trained in the same way as in the previous scenario. A similar precision for the damage field was achieved, see Fig.~\ref{fig:error_damage_nonuniform}. The U-Net has learned to estimate the global propagation of damage with the average total and pixelwise errors below 8\%, Fig.~\ref{fig:error_damage_nonuniform}. The drop in sharpness was slightly smaller compared to Scenario 1, see the example in Fig.~\ref{fig:damage_nonuniform}. 

\begin{figure}[H]
    \centering
    \includegraphics[width=0.66\textwidth]{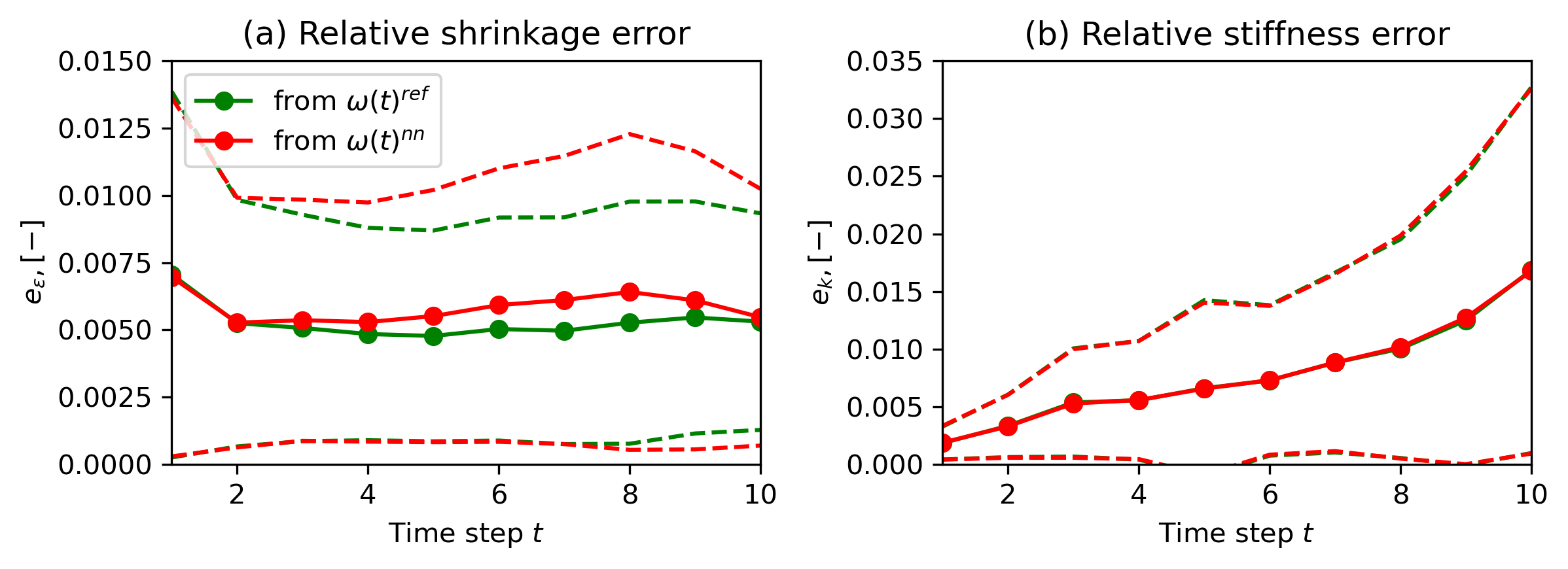}
    \caption{Mean (solid line) $\pm$ standard deviation (dashed lines) of the relative prediction errors on the test dataset, Scenario 2. The errors for the shrinkage and residual stiffness prediction are calculated from the reference damage data (green) and from the damage predicted by the U-Net (red).}
    \label{fig:error_nonuniform}
\end{figure}

\begin{figure}[H]
    \centering
    \includegraphics[width=\textwidth]{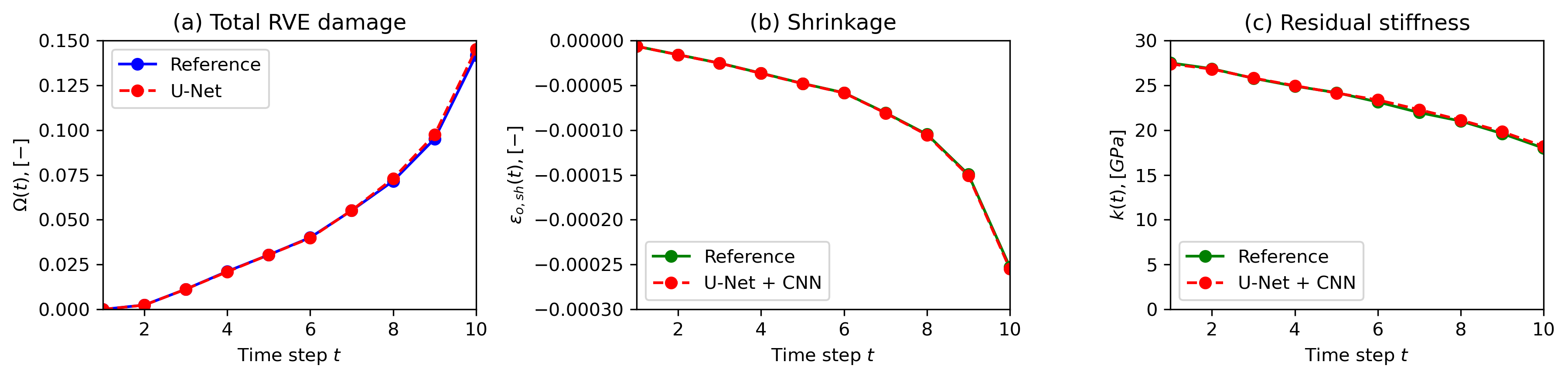}
    \caption{Prediction of the homogenised properties by the U-Net and CNN of the test geometry sample in Fig.~\ref{fig:damage_nonuniform} compared to the reference values, Scenario 2.}
    \label{fig:homog_nonuniform}
\end{figure}

The CNN predictions of the observed shrinkage and residual stiffness resulted in an error that was nearly 50\% smaller than the error in Scenario 1. Specifically, the average errors were below 1\% for shrinkage and 2\% for residual stiffness at all time steps, as illustrated in Figs.~\ref{fig:error_nonuniform}~and~\ref{fig:homog_nonuniform}. Additionally, the predictions remain consistent despite the U-Net error.

\section{Data Exploration}\label{sec:big_data}

The primary motivation for developing our deep-learning surrogate model for time-intensive FE simulations was to facilitate the analysis of how aggregate geometry affects effective shrinkage and the associated reduction in stiffness. While detailed investigations are beyond the scope of this paper and will be reserved for future research, we present a brief analysis enabled by the proposed surrogate model.

To gain insight into the influence of the microstructure on damage, effective shrinkage and stiffness, we generated
a large dataset of \numprint{100,000} microstructures and evaluated them using the neural networks presented in Section~\ref{sec:architecture}. We focus primarily on the features of the aggregates that could be realistically influenced during concrete mixing, such as the ratio of the particle sizes and their roundness. Again, we investigate the two shrinkage scenarios described in Section~\ref{sec:model}.

\subsection{Scenario 1. Uniform shrinkage}
\begin{figure}[h]
    \centering
    \includegraphics[width=\linewidth]{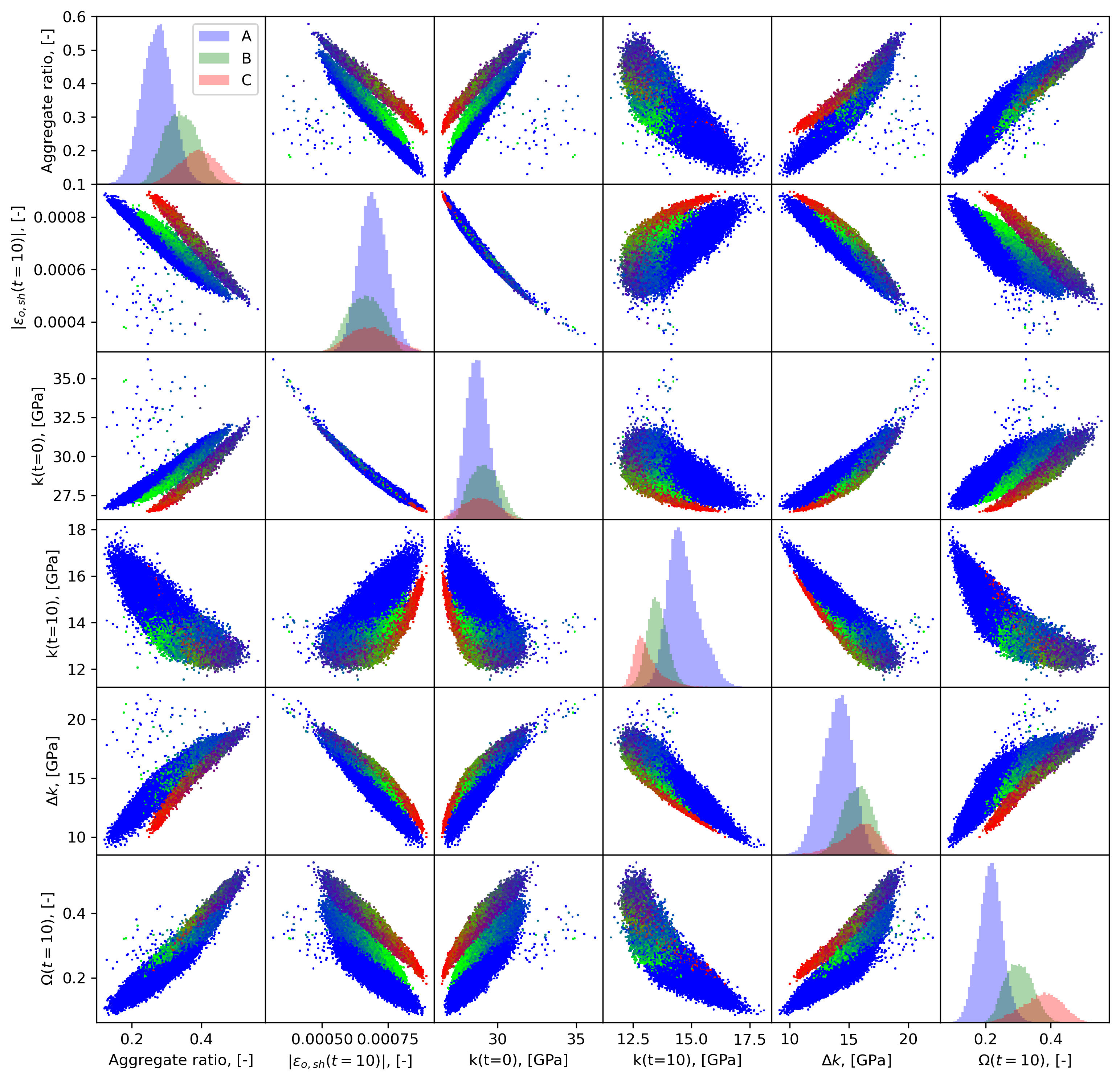}
    \caption{NN-evaluated large dataset, Scenario 1. Each data point in the subdiagonal graphs is plotted according to the RGB colour scheme in Fig.~\ref{fig:RGB}. The data on the histogram plot is split into three clusters: cluster A contains microstructures with only large particles (max. diameter = 16 mm); cluster B contains microstructures with a mix of large and medium (max. diameter = 8 mm) particles; cluster C contains all microstructures with a non-zero volume fraction of the smallest particles (max. diameter = 4 mm).}
    \label{fig:big_data_uniform}
\end{figure}

\begin{figure}
\centering
\begin{minipage}{.45\textwidth}
  \centering
  \includegraphics[width=1\linewidth]{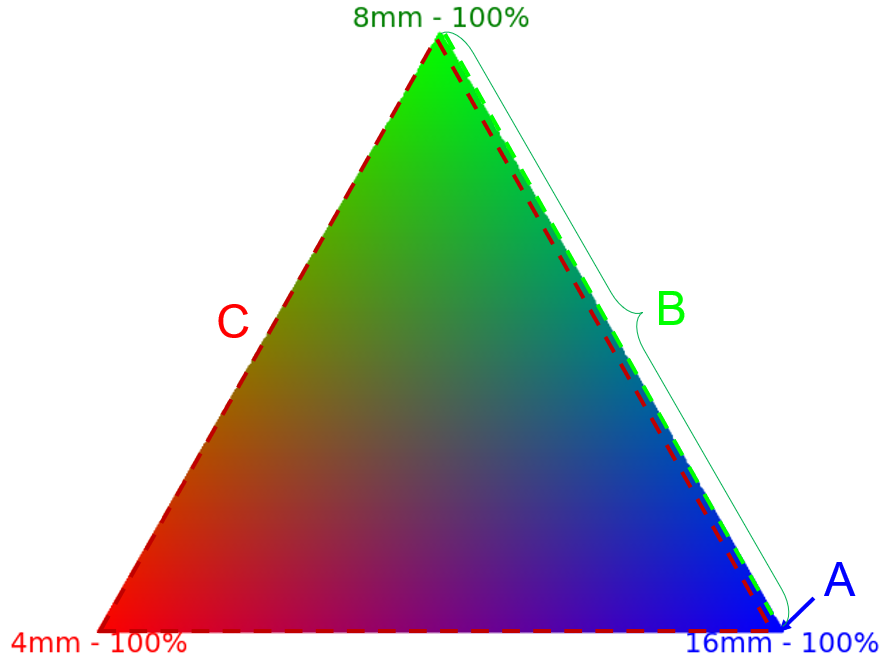}
  \captionof{figure}{RGB scheme for the data points in Figs.~\ref{fig:big_data_uniform} and \ref{fig:big_data_nonuniform}. Each channel represents the percentage of the total aggregate area belonging to a particular particle size. The resulting clusters are shown in dashed regions.}
  \label{fig:RGB}
\end{minipage}%
\hfill
\begin{minipage}{.45\textwidth}
  \centering
  \includegraphics[width=1\linewidth]{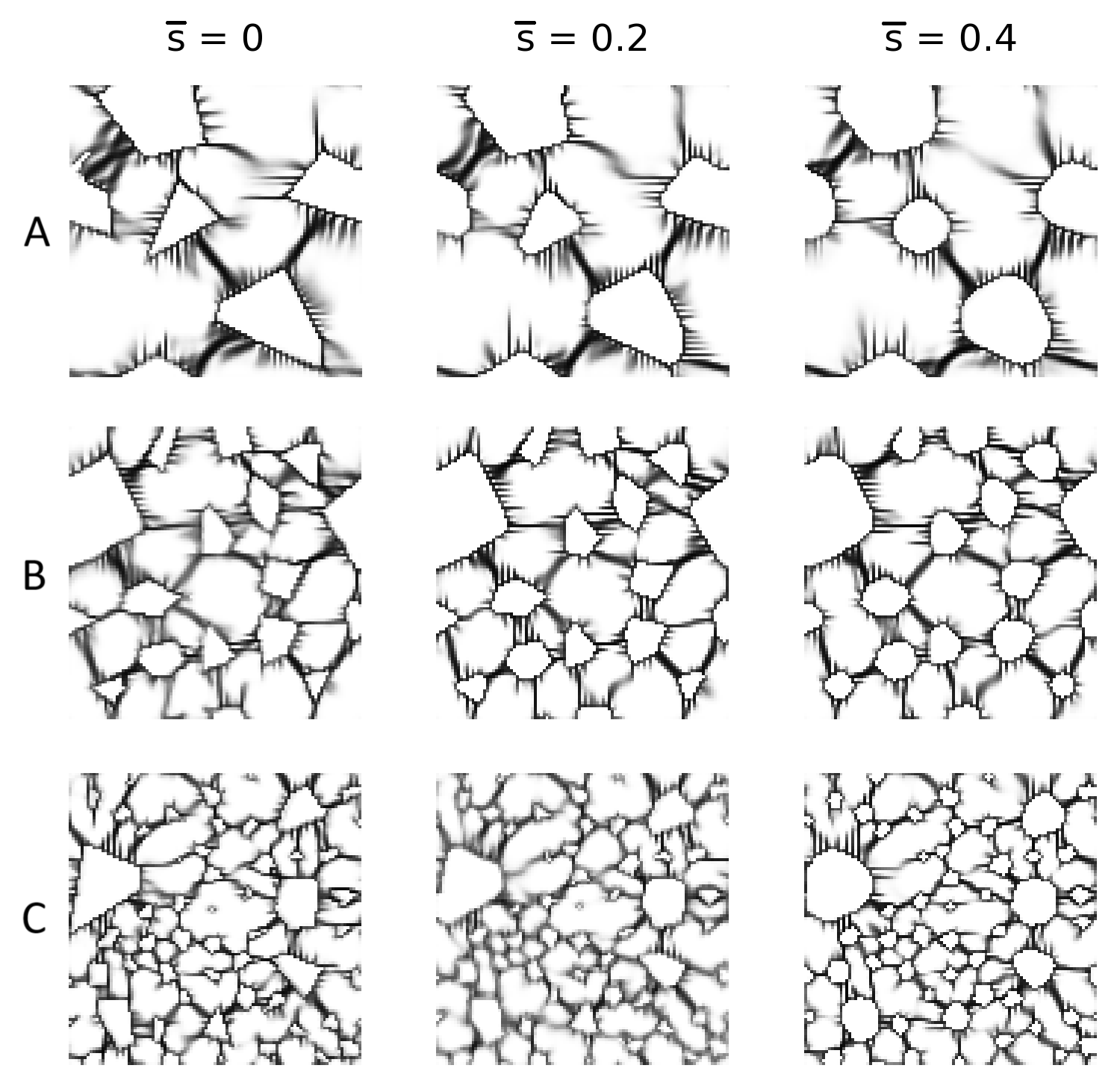}
  \captionof{figure}{Microstructure examples, clusters A, B and C, and smoothing effect according to the smoothing kernel $\overline s$.}
  \label{fig:examples_smooth}
\end{minipage}
\end{figure}

When the features of the dataset, such as the ratio of the total aggregate area to the RVE area, total damage, homogenised absolute shrinkage, initial and the final residual stiffness are plotted against each other (Fig.~\ref{fig:big_data_uniform}), three clusters can be observed. Cluster A contains microstructures with only large particles (max. diameter = 16 mm); cluster B contains microstructures with a mix of large and medium (max. diameter = 8 mm) particles; cluster C contains all microstructures with a non-zero volume fraction of the smallest particles (max. diameter = 4 mm).
To make the clusters even more apparent, we colour-coded each datapoint using the RGB scheme in which each colour channel reflects the percentage of area belonging to a particular particle size, see Fig.~\ref{fig:RGB}. 

Figure~\ref{fig:big_data_uniform} reveals that, generally, with the increasing total aggregate area, the initial stiffness and the relative total damage increase, while the observed shrinkage and final stiffness decrease. The initial stiffness and the observed shrinkage are strongly correlated, where samples with higher stiffness experience lower shrinkage. However, samples with a higher initial stiffness also tend to have a higher stiffness loss, i.e., the difference between the initial and the final residual stiffness at the final time step. 
For the same total aggregate area, cluster A tends to have lower shrinkage and higher initial and final stiffnesses than B and C, while cluster C has the lowest of the three. Although the initial stiffness is similarly distributed in all clusters, cluster A has the highest average final stiffness, the lowest average stiffness loss, as well as the lowest average total damage. 

Another feature of interest, apart from the aggregate mix (also known as the grading curve, \cite{Coenen2022}), is the shape of particles. This could represent different sources of the aggregate in real life, such as crushed stone from quarries or river gravel \citep{EN_concrete}.
To investigate the effect of the roundness of the aggregate, we randomly selected 2,000 microstructures and radially smoothed them to a varying degree by using the level-set approach \citep{Sonon2012AUL, DOSKAR2020}. Instead of taking the zeroth level of the distance to the nearest particle boundary, we considered a positive threshold $s$ for defining the contours of the aggregates. The threshold of the smoothing $s$ was set using the relative threshold $\overline{s} \in [0, 0.5]$ with respect to the radius of the particle's circumscribed circle. The particles were then centrally shrunk to keep the total aggregate area consistent. The smoothing effect is demonstrated in Fig.~\ref{fig:examples_smooth}.

\begin{figure}[h]
    \centering
    \begin{subfigure}[b]{\textwidth}
        \includegraphics[width=\linewidth]{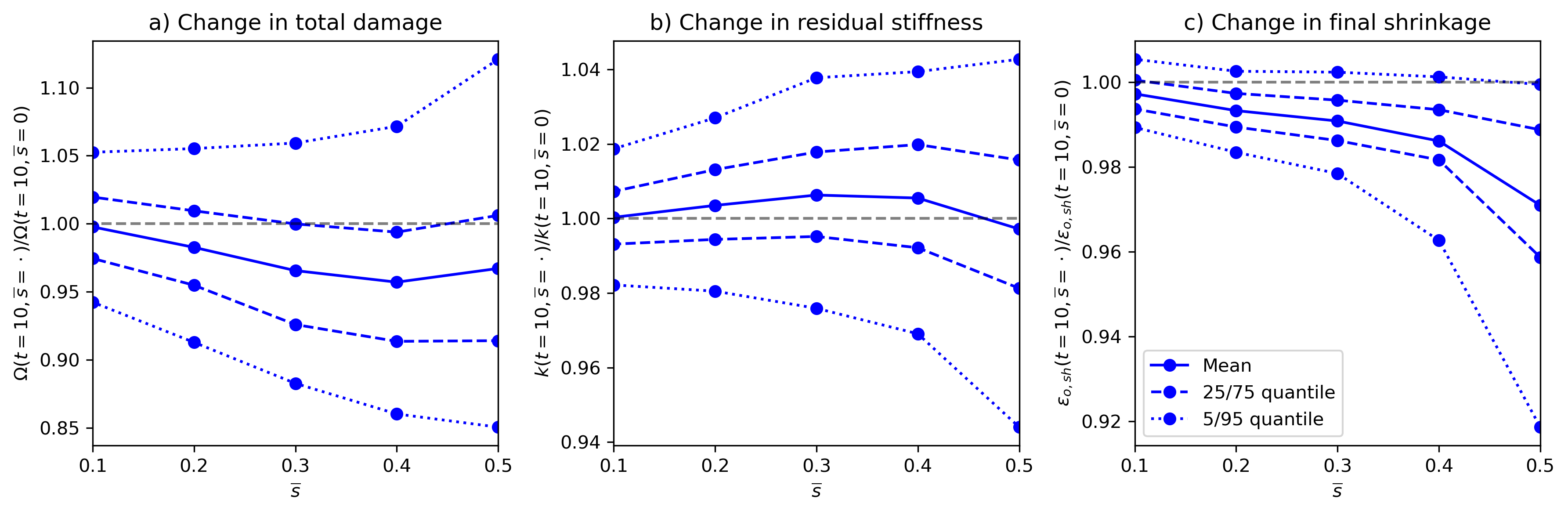}
        \caption{Dataset of microstructures of large particles (cluster A), uniform shrinkage.}
        \label{fig:smooth_a}
    \end{subfigure}
    \begin{subfigure}[b]{\textwidth}
        \includegraphics[width=\linewidth]{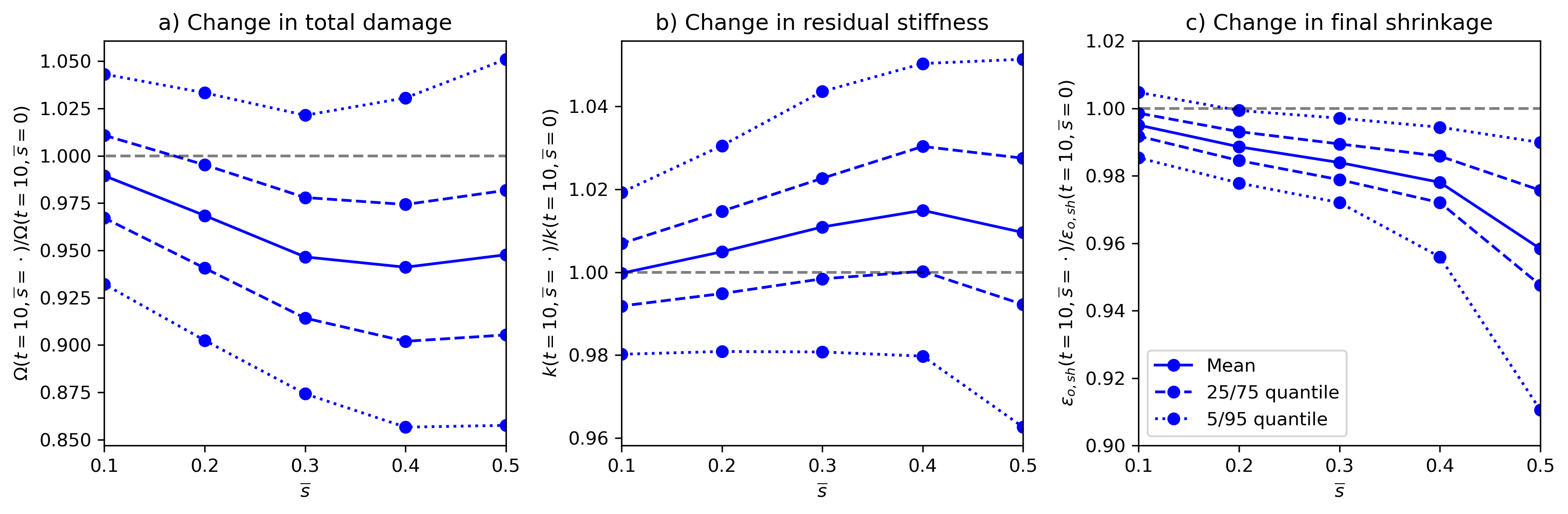}
        \caption{Dataset of microstructures of large and medium particles (cluster B), uniform shrinkage.}
        \label{fig:smooth_b}
    \end{subfigure}
    \begin{subfigure}[b]{\textwidth}
        \includegraphics[width=\linewidth]{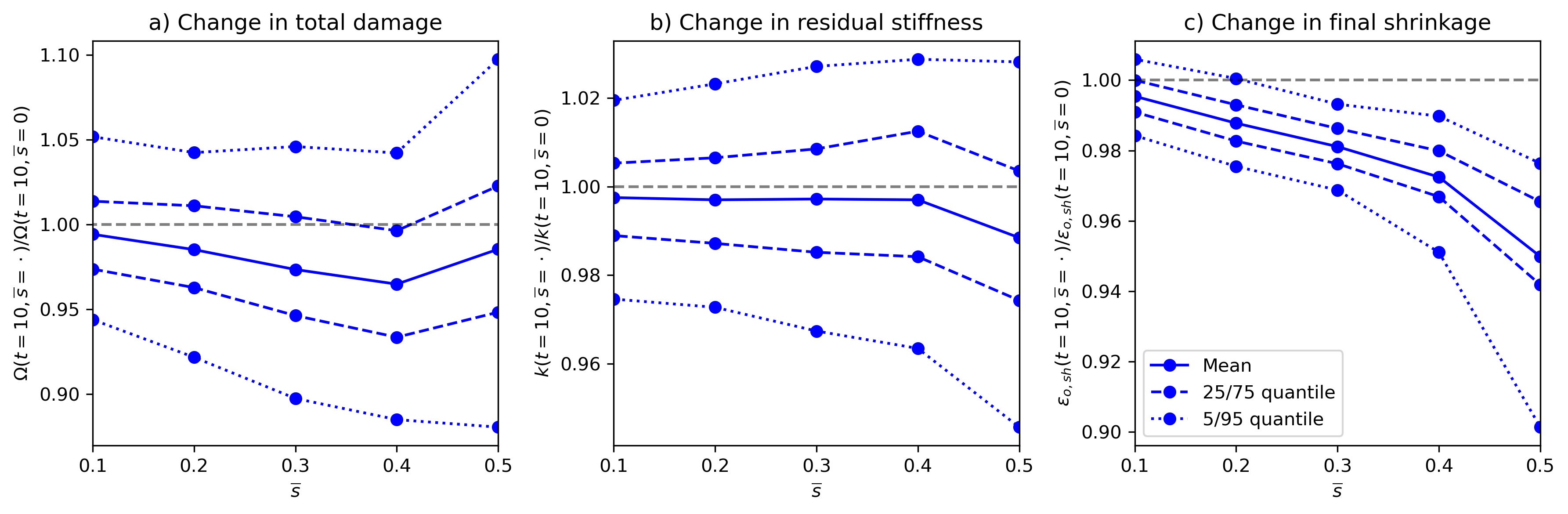}
        \caption{Dataset of microstructures with all particle types present (cluster C), uniform shrinkage.}
        \label{fig:smooth_c}
    \end{subfigure}
    \caption{Particle smoothing effect for geometries from different datasets.}
\end{figure}

When the modified microstructures are reevaluated with our surrogate model, most samples exhibit a moderate decrease in damage as the smoothing parameter $\overline{s}$ increases, with the most pronounced effect observed at $\overline{s} = 0.4$ (see Figs.~\ref{fig:smooth_a}-\ref{fig:smooth_c}, full and dashed lines), and the effect is most pronounced in the cluster B with large and medium particles, where damage decreased by up to 5\% on average. The impact on residual stiffness was not significant across any of the clusters; however, cluster B showed the strongest average increase of approximately 1.5\% at $\overline{s} = 0.4$ (Fig.~\ref{fig:smooth_b}). All clusters demonstrate observable reduction in shrinkage, but the most significant effect is noted in cluster C, where at least 95\% of samples with $\overline{s}>0.1$ exhibit some decrease in shrinkage, with the maximum average decrease of up to 4.5\%. In addition, more than 95\% of samples from clusters B and C with a smoothing kernel of $s \geq 0.2$ also show a decrease in shrinkage.

Since the observed effects are of similar magnitude to the errors of the neural networks, further investigation into the impact of particle shape is required, which we reserve for future work. 

\FloatBarrier

\FloatBarrier
\subsection{Scenario 2. Non-uniform shrinkage}

\begin{figure}
    \centering
    \includegraphics[width=\linewidth]{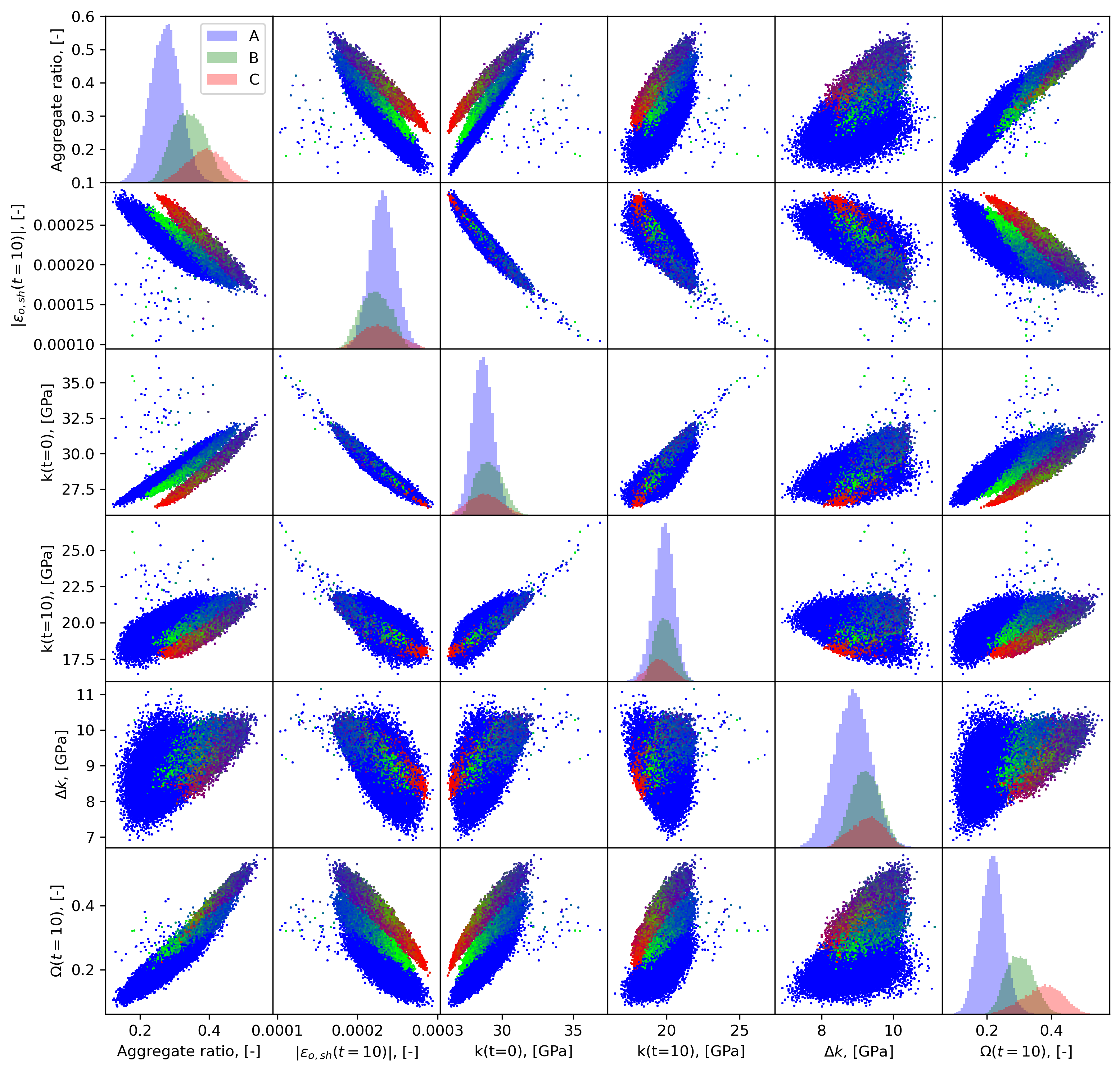}
    \caption{NN-evaluated large dataset, non-uniform shrinkage. Each data point in the subdiagonal graphs is plotted according to the RGB colour scheme in Fig.~\ref{fig:RGB}. The data on the histogram plot is split into three clusters: A) microstructures with only large particles (max. diameter = 16 mm), B) microstructures with a mix of large and medium (max. diameter = 8 mm) particles, and C) microstructures with all types of particles present, including the smallest (max. diameter = 4 mm).}
    \label{fig:big_data_nonuniform}
\end{figure}

The same microstructure dataset was evaluated within the non-uniform shrinkage scenario, as shown in Fig.~\ref{fig:big_data_nonuniform}. Most of the same trends with regard to clustering can be observed. For instance, the total area of particles, the initial stiffness, and the relative total damage increase are all positively correlated. However, final stiffness is now positively correlated with the initial stiffness, and the clusters experience similar stiffness loss and have similar distributions of the final stiffness. 
For the same total aggregate area, cluster A again tends to have lower absolute shrinkage and damage than clusters B and C. 

Since this dataset simulates the behaviour of the outer layer of a drying concrete beam, we are interested in the influence of the aggregate distribution in the matrix on the observed shrinkage. In particular, larger particles typically do not occur at the surface of the concrete, so the distribution of the aggregate in a cast concrete varies between the surface and deeper sections. On the other hand, the normal dispersion scenario could represent a concrete cast that has been cut after the initial solidifying phase. To investigate this effect, we removed all aggregates in the lower (more exposed to drying) layer of the microstructures in the large dataset, randomly selecting vertical areas of 5, 10, 15, 25, or 30 pixels, and let the neural networks predict the damage and homogenized properties on the new data.

\begin{figure}[h]
    \centering
    \includegraphics[width=\textwidth]{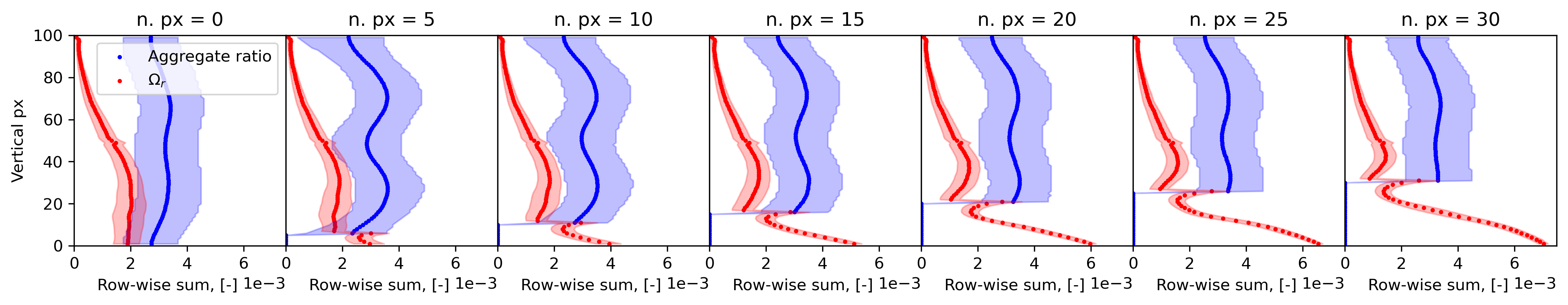}
    \caption{Average row-wise damage profile depending on the thickness of the aggregateless outer layer in pixels.}
    \label{fig:vertical_damage}
\end{figure}

\begin{figure}[h]
    \centering
    \includegraphics[width=\textwidth]{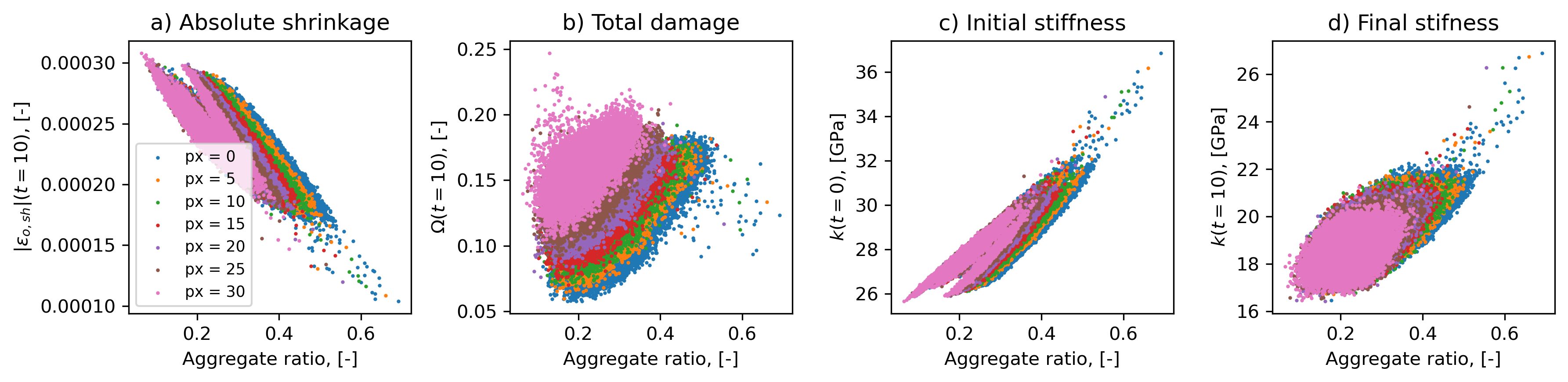}
    \caption{Comparison of the datasets with different cover layer widths for the overall damage, shrinkage, and final residual stiffness.}
    \label{fig:damage_effect_scatter}
\end{figure}

\begin{figure}[h]
    \centering
    \includegraphics[width=\textwidth]{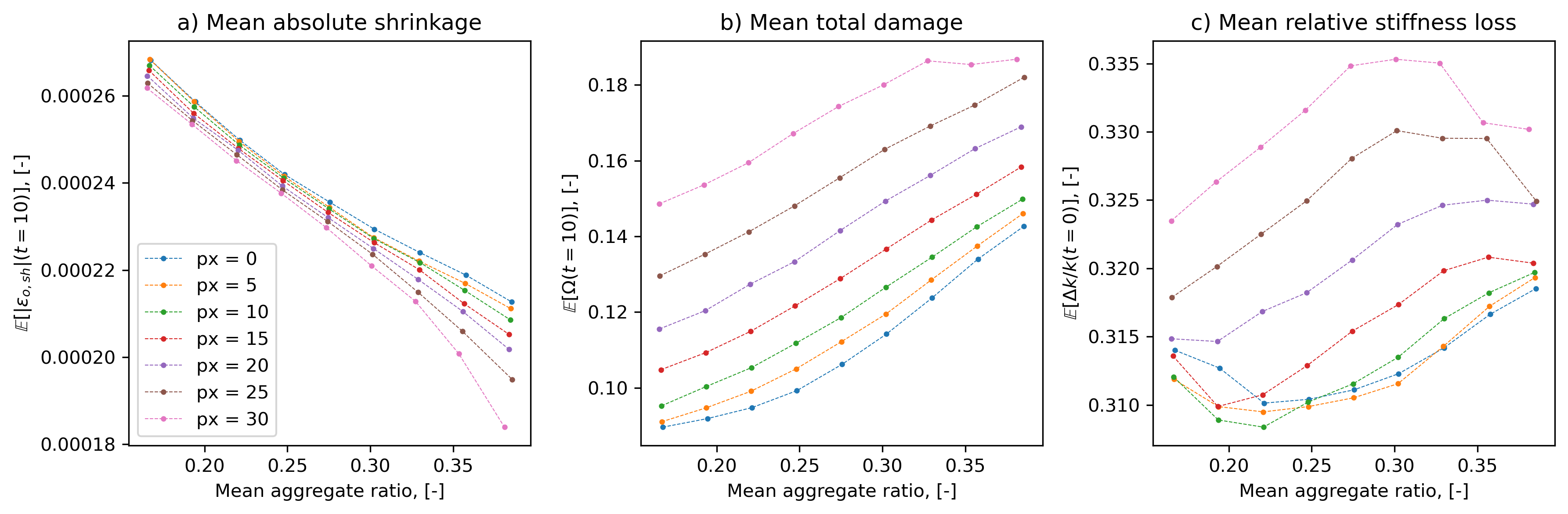}
    \caption{Comparison of the average ($\mathbb E (\cdot)$) final damage, shrinkage, and stiffness loss of the initial dataset and the datasets with the erased aggregate layers.}
    \label{fig:damage_effect}
\end{figure}

Corresponding results are shown in Fig.~\ref{fig:vertical_damage}, where we averaged the row-wise sum of final damage, $\Omega_r(t=10)$. We can see that more damage accumulates in the horizontal layers of these microstructures that are closer to the surface (on the bottom). With increasing height of the outer layer, damage tends to redistribute towards the outer layer, i.e., decrease slightly in the upper area where the particles are present, and increase significantly in the outer layer. Since the data modification decreases the total aggregate area, the samples from the original dataset cannot be compared to the samples of the new dataset in a one-to-one fashion, as the aggregate area itself is a strong effective property predictor, as seen in Fig.~\ref{fig:damage_effect_scatter}. To decouple the change in particle distribution from the change in the total aggregate area, we group both datasets by the ranges of aggregate area. This way, we can compare the samples of the modified dataset to the samples of the initial dataset that have a similar total aggregate area, see Fig.~\ref{fig:damage_effect}. The absolute final observed shrinkage decreases slightly with the wider aggregateless outer layer. The effect is the strongest for the aggregate area ratio greater than 30\%. The most pronounced effect is on the damage, where samples with wider outer layers have higher total final damage, which is consistent with the findings in Fig.~\ref{fig:vertical_damage}. Only samples with the layer height larger than 20 pixels ($\approx 6$~mm) show a significant increase in the relative stiffness loss, and the overall effect is low. The correlation between an increase in damage and a decrease in shrinkage supports the previous analysis in Fig.~\ref{fig:big_data_nonuniform}.

\FloatBarrier
\section{Conclusion}\label{sec:conclusion}

This paper presented a deep learning surrogate model for predicting the evolution of shrinkage-induced damage in concrete. The approach combined an auto-regressive U-Net for full-field scalar damage prediction with a convolutional neural network (CNN) that estimated homogenised mechanical responses, such as residual stiffness and observed shrinkage. Trained on large-scale synthetic datasets generated via nonlinear mesoscale simulations, the model reproduced full-field damage evolution and homogenised responses with a typical error of 1 to 5\%. The results demonstrate that complex multiscale behaviour can be captured at a fraction of the computational cost, without requiring detailed physical simulations for every new microstructure.

To illustrate its potential, we used the surrogate for a preliminary statistical analysis of how aggregate topology affects shrinkage-induced damage. Leveraging datasets of thousands of varied microstructures, the model enabled rapid evaluation of trends related to particle size, shape, and distribution. 

All findings should be interpreted in the context of a deliberately idealised 2D framework. While these observations are necessarily shaped by the simplified two-dimensional setting and synthetic data, the analysis exemplifies the feasibility of a data-driven, large-scale exploration of microstructure–response relationships in cementitious materials. Although the surrogate demonstrates good generalisation to unseen microstructures, the conclusions are not meant to be definitive. Rather, this work serves as a proof of concept showing that full-field prediction and homogenised response estimation can be effectively learned from simulation data. Extending the model to three-dimensional settings would be a natural next step toward realistic structural-scale applications. In addition, the use of more sophisticated microstructural descriptors~\citep{PYRZ1994197, XU2014206} could provide further insight into the link between geometry and mechanical response, enhancing both the interpretability and generalizability of surrogate models.

Implementation and trained models are available on \href{https://github.com/LiyaGaynutdinova/Auto-Regressive-Unet-for-Concrete-Damage}{GitHub}, with accompanying datasets accessible via Zenodo.

\FloatBarrier

\section*{CRediT Author Statement}
LG: Methodology, Software -- Deep Neural Networks, Investigation, Writing -- Sections \ref{sec:architecture}, \ref{sec:experiments}, \ref{sec:big_data}, Visualization; 
PH: Methodology, Conceptualization, Software -- Concrete Damage Simulation, Writing -- Section \ref{sec:model}, Review \& Editing; 
OR: Methodology, Writing -- Review \& Editing, Supervision;
FH: Methodology -- Deep Neural Networks, Writing -- Review \& Editing;
MD: Methodology, Software -- Level Set, Writing -- Review \& Editing, Supervision.

\section*{Acknowledgments}
PH gratefully acknowledges the financial support provided by the Czech Science Foundation under project No.~25-17557J.
MD's work was supported by the Czech Science Foundation through project No.~24-11845S.

\bibliographystyle{elsarticle-harv} 
\bibliography{cas-refs}

\end{document}